\pdfoutput=1
\documentclass[11pt]{article}

\usepackage[a4paper,margin=2.5cm]{geometry}
\usepackage[T1]{fontenc}
\usepackage{graphicx}
\usepackage{framed,multirow}
\usepackage{float}

\usepackage{amssymb}
\usepackage{latexsym}
\usepackage{algorithm}
\usepackage{algpseudocode}
\usepackage{url}

\usepackage{booktabs}
\usepackage{amsmath}
\usepackage{bm}
\usepackage{comment}
\usepackage[dvipsnames]{xcolor}

\usepackage[numbers,sort&compress]{natbib}
\usepackage[colorlinks=true,linkcolor=blue,citecolor=blue,urlcolor=blue]{hyperref}
\usepackage{authblk}

\title{Knowledge-Constrained Shape Optimization with a Mixture-of-Experts Neural Operator for High-Confidence Design}

\author[1,2,3]{Wenhao Fan}
\author[2,3]{Yuanwei Bin\thanks{Corresponding author. E-mail: ybin@eitech.edu.cn}}
\author[3]{Jianghan Gu}
\author[4]{Wenfa Luo}
\author[4]{Jiao Xiang}
\author[2]{Yuntian Chen}
\author[2]{Shiyi Chen}

\affil[1]{School of Ocean and Civil Engineering, Shanghai Jiao Tong University, Shanghai 200240, China}
\affil[2]{Eastern Institute of Technology, Ningbo, Ningbo 315200, Zhejiang, China}
\affil[3]{TenFong Technology Co., Ltd., Shenzhen 518000, Guangdong, China}
\affil[4]{IM Motors Technology Co., Ltd., Shanghai 201210, China}

\date{}

\begin{document}

\maketitle

\begin{abstract}
Engineering shape optimization presents challenges at both the application and technical levels.
At the application level, the specification of optimization settings, including editable regions, deformation ranges, and design-preservation constraints, relies heavily on domain expertise and is typically performed manually by experienced engineers.
At the technical level, surrogate-based optimization methods often exhibit limited reliability when applied to heterogeneous geometry databases or when optimization trajectories enter out-of-distribution regions.
To address these challenges, we propose a knowledge-constrained shape-optimization framework that translates knowledge-based constraints and user intent into quantifiable parameters of DFFD-based deformation operators, thereby enabling engineering-aware and controllable constrained optimization.
We further develop a Mixture-of-Experts Neural Operator (MoE-NO) to enhance surrogate modeling performance over heterogeneous aerodynamic datasets.
Compared with baseline models, including Transolver \cite{wu2024transolver} and DragSolver \cite{liu2025dragsolver}, MoE-NO achieves improved drag-prediction accuracy and enhanced consistency in trend prediction.
Based on the MoE-NO encoder and the Mahalanobis distance, we introduce an uncertainty-estimation approach to identify out-of-distribution geometries during the optimization process.
Design candidates associated with high uncertainty are selectively evaluated using a physics-based solver for local sample enrichment, thereby improving optimization reliability without incurring the cost of evaluating all candidates.
Experiments conducted on in-house MPV, SUV, and Sedan datasets demonstrate that MoE-NO achieves a test-set MAPE of $1.16\%$, outperforming the best baseline result of $1.52\%$.
In addition, MoE-NO improves trend-prediction accuracy to $94.34\%$, compared with the best baseline accuracy of $90.34\%$.
Vehicle shape-optimization experiments yield CFD-validated drag coefficient reductions in the range of approximately $4\%$ to $10\%$.
\end{abstract}

\vspace{0.5em}
\noindent\textbf{Keywords:} Shape optimization; Knowledge-constrained design; Surrogate-assisted optimization; Mixture-of-Experts Neural Operator

%\linenumbers

%% Main text
\section{Introduction}

Physics-governed shape optimization seeks geometry modifications that improve a performance objective evaluated by a numerical physics model.
It appears in aerodynamic design~\cite{jameson1988aerodynamic,martins2022aerodynamic,skinner2018state,yuan2026resolvent}, thermal design~\cite{park2004numerical}, structural design~\cite{wall2008isogeometric}, and multidisciplinary design~\cite{samareh2001survey,martins2021engineering}, and is usually formulated through geometry parameterization~\cite{samareh2001survey,wei2024deepgeo}, high-fidelity simulation~\cite{moukalled2015finite,versteeg1995computational}, adjoint or derivative-free search~\cite{jameson1988aerodynamic,jones1998efficient,liu2025asynchronous}, surrogate modeling~\cite{umetani2018learning,li2024machine}, and multidisciplinary design optimization~\cite{martins2021engineering}.
In a typical workflow, a baseline geometry is mapped to a finite-dimensional design space, candidate geometries are evaluated by a physics solver, and the design variables are updated until a performance criterion is improved.
This formulation has become a standard computational route for inverse design~\cite{chen2026optimization,sun2026geometry}, but its use in industrial geometry optimization remains limited by two practical issues.

The first challenge is the construction of engineering constraints.
In many design problems, admissible modifications are not specified only by analytic inequalities.
They are also encoded in design standards~\cite{crandall2002designing,hu2024association}, engineering guidelines~\cite{hucho2013aerodynamics,schuetz2015aerodynamics,barnard2001road}, benchmark studies~\cite{ahmed1984some,heft2012introduction,ekman2020assessment}, internal requirements, and expert experience~\cite{chapman2001application,la2012knowledge,kuegler2023evolution}.
Knowledge-based engineering sources~\cite{chapman2001application,la2012knowledge,kuegler2023evolution} specify which regions may be edited, which deformation directions are admissible, what deformation bounds should be imposed, and which properties must be preserved, such as packaging, interfaces, smoothness, manufacturability, and design identity.
Conventional optimization workflows usually rely on engineers to translate this knowledge into design variables and bounds before the numerical search starts.
As a result, admissible deformation space is difficult to construct automatically and remains weakly coupled with the subsequent solver-based optimization.
More recently, large language model (LLM) based multi-agent systems have shown promise in automating engineering analysis and simulation workflows, such as agentic CFD case generation~\cite{yue2025foamgpt}, end-to-end finite element analysis for solid mechanics~\cite{sarker2026multi}, and the aerodynamic design of vehicles, airfoils, and wings~\cite{liu2026aeroagent,fang2026agentic,fan2026airfoilagent,elrefaie2025ai,lee2025aerodynamic,jin2025closed}, which motivates coupling knowledge interpretation more tightly with the subsequent solver-based optimization.
The second challenge is the cost of repeated high-fidelity solver evaluations.
Each optimization iteration may require geometry modification, mesh generation or mesh deformation, numerical solving, and post-processing.
For high-fidelity simulations, such as CFD evaluations~\cite{moukalled2015finite,versteeg1995computational} of external-flow problems~\cite{ekman2020assessment,qin2024drivaer}, this cost can dominate the optimization loop.
%Fast meshing~\cite{kanade2009rapid} and GPU-accelerated simulation~\cite{niedermeier2018ultra} reduce the cost of individual evaluations, but they do not remove the need to limit the number of high-fidelity solver calls.
This has motivated surrogate-assisted optimization~\cite{jones1998efficient}, where a learned model~\cite{umetani2018learning,li2024machine} approximates the solver response and is used to screen candidate designs before expensive validation.

Neural operators provide a useful class of surrogates for this purpose.
Methods such as DeepONet~\cite{lu2021learning} and Fourier neural operator~\cite{li2020fourier} learn mappings between function spaces and have been used to accelerate parametric physical simulations,increasingly supported by large-scale aerodynamic datasets that span diverse flow regimes~\cite{kanchi2026unifoil}.
Related data-driven models have also been developed for unsteady aerodynamics and aeroelasticity~\cite{kou2021data,zhang2026new} and for high-fidelity flow-field prediction over lifting surfaces~\cite{cui2026high}.
For geometry-dependent problems, point-cloud encoders~\cite{zhao2021point,wu2024point} and transformer-based representations for automotive drag prediction~\cite{liu2025dragsolver,qi2026geometry,gu2026geoformer,zhang2026reto,yang2025spatially} further allow physical quantities to be predicted from complex three-dimensional shapes.
When used inside an optimization loop, these models can reduce the number of direct calls to the high-fidelity solver.
However, their accuracy depends strongly on the distribution and coverage of the solver-labeled data used for training.

\begin{figure}
    \centering
    \includegraphics[width=0.85\textwidth] {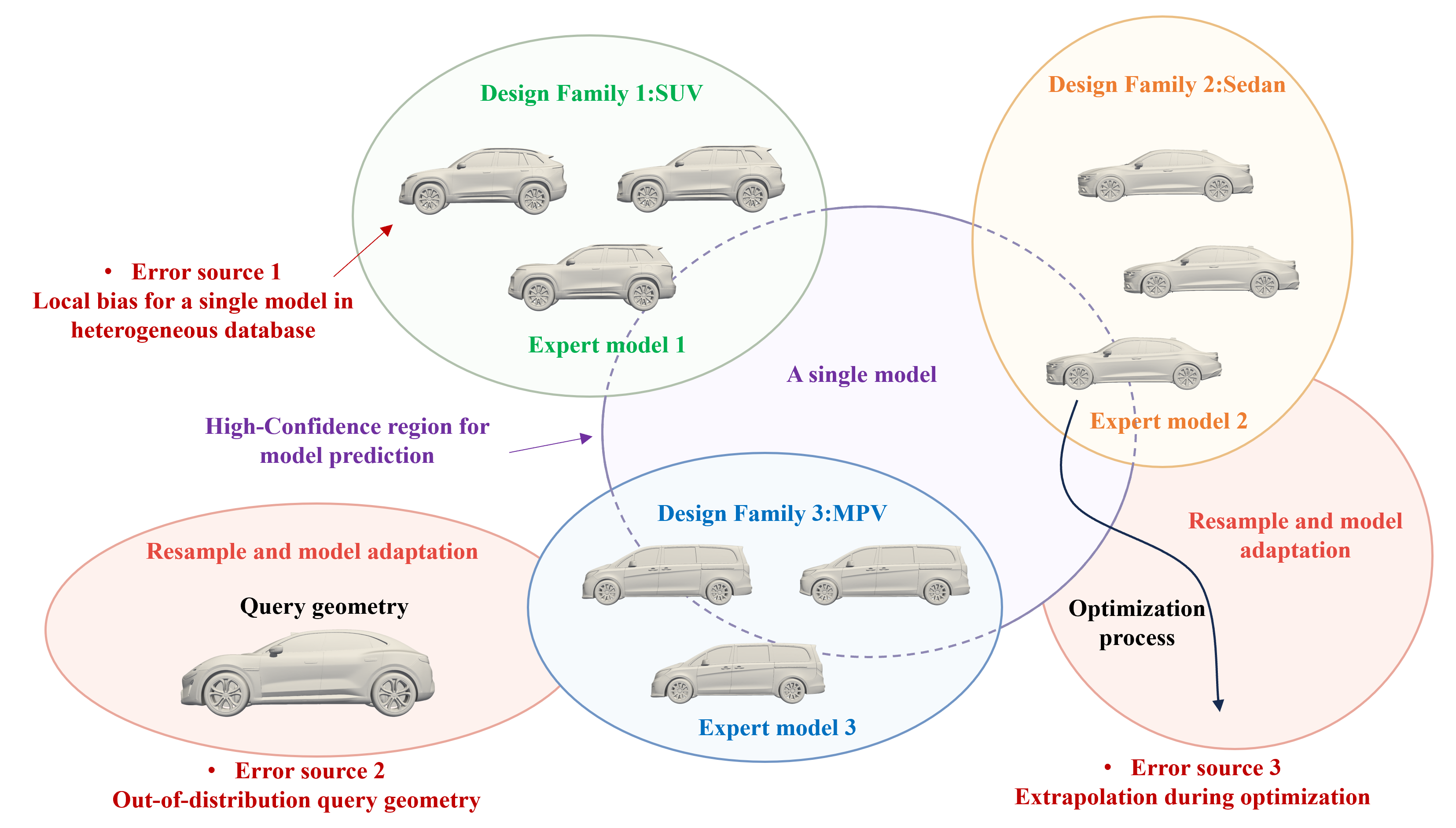}
    \caption{Three major sources of error in  surrogate-assisted inverse design.}
    \label{fig:error_ssource}
\end{figure}

This dependency creates three major sources of error in surrogate-assisted inverse design, as shown in Fig.~\ref{fig:error_ssource}.
First, historical simulation databases are often heterogeneous.
They may contain multiple design families, operating regimes, or local geometric variants, rather than samples from a single distribution.
When such data are represented by one global surrogate~\cite{li2024machine}, local bias can arise because a single model must fit several distinct geometry--performance relations~\cite{umetani2018learning}.
Second, the query geometry may be poorly covered by the historical simulation database~\cite{pimentel2014review,chandola2009anomaly}.
In this case, even before optimization starts, the surrogate may be asked to predict outside the region where it has reliable training support.
Third, the optimization process itself can induce extrapolation.
Even if the initial geometry is in distribution, the optimizer may move candidate designs into sparsely sampled regions of the design space~\cite{kaya2019deep}.
These three cases can lead to unreliable surrogate predictions~\cite{jones1998efficient,li2024machine} and may produce candidate optima that are not confirmed by the physics solver.

This work develops a knowledge-constrained shape optimization method with the MoE-NO model.
The method first addresses the constraint-construction challenge by converting domain knowledge $\mathcal{K}$ into editable control boxes $\mathcal{C}$, an admissible deformation space $\Omega(\mathcal{C})$, and preservation constraints $\mathcal{H}$ through Retrieval-Augmented Generation (RAG) over task-relevant design rules and vision-language-model (VLM)-based grounding of editable geometric regions~\cite{lewis2020retrieval,gao2023retrieval,radford2021learning}.
A Direct Free-Form Deformation (DFFD) operator~\cite{sederberg1986free,hsu1992direct,lamousin1994nurbs} then converts the grounded regions, admissible directions, and deformation bounds into smooth local geometry updates~\cite{jakobsson2007mesh,luke2012fast,witteveen2010explicit}.
In this way, the numerical optimizer searches within a knowledge-constrained design space rather than an unconstrained perturbation space.

To enable high-confidence design, the method addresses the three surrogate-error sources using a physics-solver-in-the-loop strategy and uncertainty quantification.
To reduce the error caused by heterogeneous historical databases, the surrogate is formulated as the MoE-NO model.
A gating network assigns each geometry to multiple expert neural operators, allowing different local geometry--performance relations to be represented within one optimization surrogate~\cite{jacobs1991adaptive,jordan1994hierarchical,zhou2012ensemble}.
To reduce the error caused by insufficient database coverage, the encoder of the MoE-NO surrogate is used before optimization to test whether the query geometry is covered by the historical solver-labeled database.
If the query is out-of-distribution (OOD), local samples are generated within a slightly expanded version of the admissible deformation space $\Omega(\mathcal{C})$, evaluated by the physics solver, and added to the database for MoE-NO model adaptation.
To reduce the error caused by search-induced extrapolation, a Mahalanobis-distance-based uncertainty measure is computed in the encoder latent space during optimization.
If the surrogate optimum is uncertain, additional local solver-labeled samples are generated and the MoE-NO surrogate is updated before re-optimization.
Thus, high-fidelity evaluations are used for database enrichment and local error correction rather than for every candidate design.

The proposed framework is formulated for general physics-solver-in-the-loop shape optimization.
In this paper, the solver is instantiated as a high-fidelity CFD solver~\cite{moukalled2015finite,versteeg1995computational}, and the method is evaluated on vehicle aerodynamic optimization~\cite{hucho2013aerodynamics,schuetz2015aerodynamics,barnard2001road}.
The objective is to reduce the drag coefficient $C_d$ while preserving the design intent of production-intent geometries~\cite{elrefaie2024drivaernet,qiu2026drivaerstar}.
The automotive case provides a representative testbed because it combines constrained geometric deformation, heterogeneous historical simulation data, expensive CFD evaluation, and the need for CFD validation of optimized designs.

The remainder of this paper is organized as follows.
Section~\ref{sec:methods} presents the proposed knowledge-constrained shape optimization framework.
Section~\ref{sec:setup_details} describes the experimental setup, including data preparation, surrogate training, CFD simulation, and optimization settings.
Section~\ref{sec:results} reports the main results.
Section~\ref{sec:conclusions} concludes the paper.

\section{Methods}
\label{sec:methods}

\subsection{Problem formulation}
\label{subsec:general_problem_formulation}

We formulate the task as knowledge-constrained shape optimization for an engineering geometry whose original functional and visual intent must be preserved.
The input is a baseline geometry $G$ from a specific design family, together with multi-view observations, domain knowledge, and a historical simulation database.
The goal is to improve a physics-governed performance quantity while restricting geometric changes to regions and deformation ranges permitted by the available knowledge.
In this work, knowledge refers to design standards, engineering guidelines, benchmark studies, internal requirements, and task-specific aerodynamic principles.
These knowledge sources define where the geometry may be edited, how each editable region may be deformed, and which functional or visual constraints must be preserved.
In the vehicle demonstration used in this work, $G$ is a production-intent vehicle geometry and the performance quantity is the aerodynamic drag coefficient $C_d$.

We denote the collected knowledge as $\mathcal{K}$ and summarize its role as a mapping
\begin{equation}
\mathcal{K}
\longrightarrow
\left(
\mathcal{C},
\Omega\left(\mathcal{C}\right),
\mathcal{H}
\right),
\end{equation}
where $\mathcal{C}$ denotes the fused three-dimensional editable control-box parameters, $\Omega(\mathcal{C})$ denotes the corresponding admissible deformation space and $\mathcal{H}$ denotes the constraints that preserve packaging space, safety clearance, manufacturability, component interfaces, geometric continuity, and design identity.
The optimization is therefore not performed over arbitrary shape perturbations, but over a knowledge-constrained design space derived from $\mathcal{K}$.

We denote the available historical simulation database as
\begin{equation}
\mathcal{B}_{\mathrm{hist}}
=
\left\{
\left(
G^{(i)},
J_{\mathrm{phys}}(G^{(i)})
\right)
\right\}_{i=1}^{N_d},
\end{equation}
where $G^{(i)}$ is the $i$-th historical geometry, $J_{\mathrm{phys}}(G^{(i)})$ is its high-fidelity physics-evaluated objective value, and $N_d$ is the number of historical samples.

\begin{figure}
    \centering
    \includegraphics[width=\textwidth] {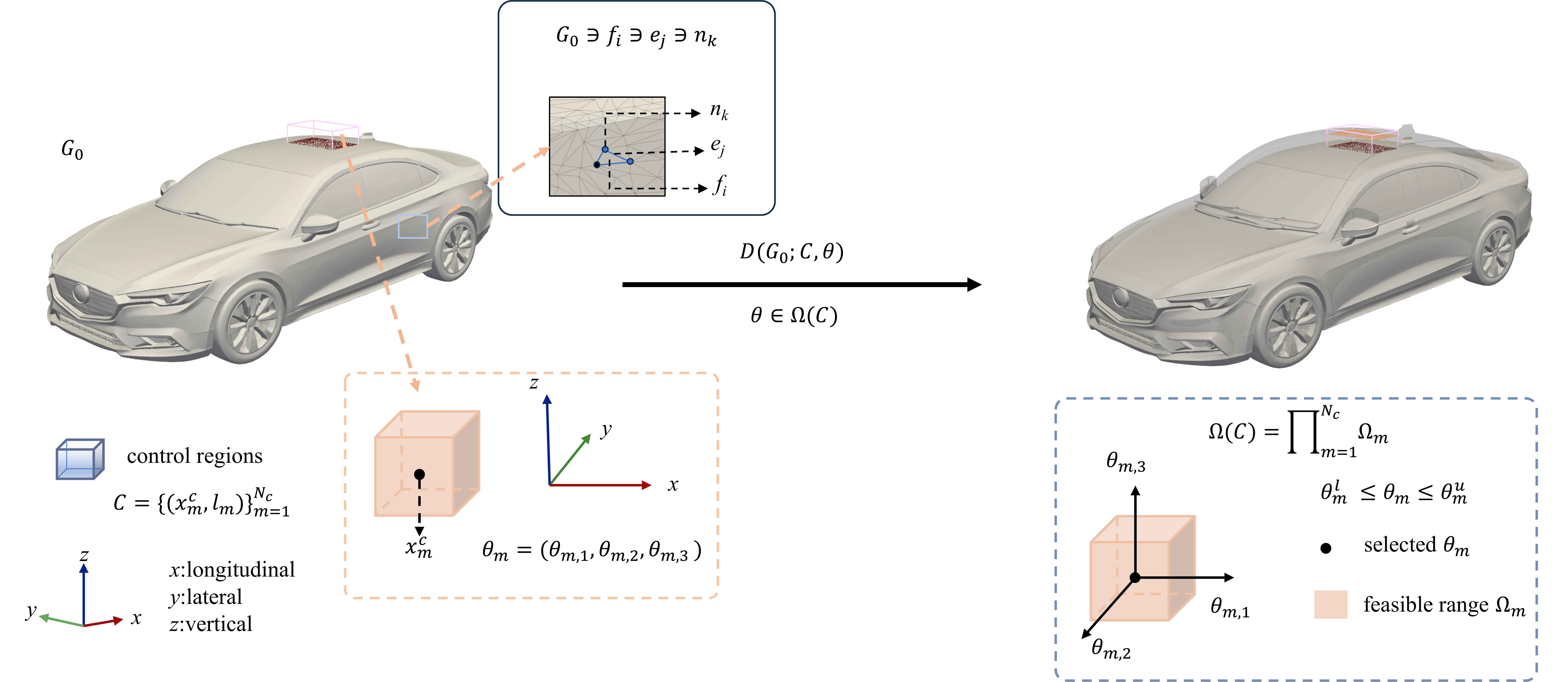}
    \caption{Knowledge-constrained design-space construction for physics-driven shape optimization.}
    \label{fig:sketch}
\end{figure}

As shown in Fig.~\ref{fig:sketch}, the baseline geometry is represented as a structured surface hierarchy,
\begin{equation}
G =
\left(
\mathcal{F},
\mathcal{E},
\mathcal{N}
\right),
\qquad
\mathcal{F}=\{\bm{f}_i\}_{i=1}^{N_f},
\quad
\mathcal{E}=\{\bm{e}_j\}_{j=1}^{N_e},
\quad
\mathcal{N}=\{\bm{n}_k\}_{k=1}^{N_n}.
\label{eq:geo}
\end{equation}
Here, $\bm{f}_i$ denotes a surface face, $\bm{e}_j$ denotes an edge, and $\bm{n}_k\in\mathbb{R}^3$ denotes a surface node with spatial coordinates.
Each face is associated with a subset of edges, and each edge is associated with its endpoint nodes.
The quantities $N_f$, $N_e$, and $N_n$ denote the total numbers of faces, edges, and nodes in the discretized geometry, respectively.
This hierarchy preserves the topological relationships required for local interpretation and constrained deformation.
For multimodal interpretation, the geometry is also rendered into orthographic multi-view images $\mathcal{I}=\{I_v\}_{v=1}^{N_v}$.
Here, $I_v$ denotes the $v$-th rendered view and $N_v$ denotes the number of rendered views.

The editable portion of the design space is also plotted in Fig.~\ref{fig:sketch}, which is parameterized by a set of localized cuboidal control boxes. 
The $m$-th control-box parameter is defined as
\begin{equation}
c_m
=
\left(
\bm{x}_m^c,\bm{l}_m
\right),
\end{equation}
where $\bm{x}_m^c\in\mathbb{R}^3$ is the box center and $\bm{l}_m\in\mathbb{R}_+^3$ is the side-length vector.
We denote the collection of editable control-box parameters as
\begin{equation}
\mathcal{C}
=
\left\{
c_m
\right\}_{m=1}^{N_c},
\label{eq:ctrl_reg}
\end{equation}
where $N_c$ is the number of editable control boxes.

Each control box is associated with a local deformation vector
\begin{equation}
\bm{\theta}_m
=
(\theta_{m,1},\theta_{m,2},\theta_{m,3})
\in\mathbb{R}^3,
\end{equation}
whose three components represent deformation amplitudes along the $x$-, $y$-, and $z$-axes, respectively.
The full set of deformation vectors is denoted by
\begin{equation}
\bm{\theta}
=
\left\{
\bm{\theta}_m
\right\}_{m=1}^{N_c}.
\label{eq:defor}
\end{equation}
The coordinate axes are defined in the object coordinate frame of the target design family. 
For the automotive case, the $x$, $y$, and $z$ axes correspond to the longitudinal, lateral, and vertical directions, respectively.

Given the initial geometry $G$, the deformed geometry is obtained through a constraint-aware deformation operator
\begin{equation}
G'(\bm{\theta})
=
\mathcal{D}(G;\mathcal{C},\bm{\theta}),
\end{equation}
where $\mathcal{D}$ applies the prescribed local deformations within the editable control boxes while preserving geometric consistency and design constraints.

For each control box, the admissible deformation range is defined as
\begin{equation}
\Omega_m(c_m)
=
\left\{
\bm{\theta}_m\in\mathbb{R}^3
\mid
\bm{\theta}_m^l
\preceq
\bm{\theta}_m
\preceq
\bm{\theta}_m^u
\right\},
\end{equation}
where $\bm{\theta}_m^l$ and $\bm{\theta}_m^u$ are the lower and upper deformation bounds retrieved from domain rules by the RAG module for the $m$-th control box.
The three components of these bounds describe the admissible deformation range along the three orthogonal coordinates.
The overall admissible deformation space is then given by
\begin{equation}
\Omega(\mathcal{C})
=
\prod_{m=1}^{N_c}
\Omega_m(c_m).
\end{equation}

The resulting physics-driven design optimization problem is formulated as
\begin{equation}
\bm{\theta}^{*}
=
\arg\min_{\bm{\theta}\in\Omega(\mathcal{C})}
J_{\mathrm{phys}}
\left(
G'(\bm{\theta})
\right)
=
\arg\min_{\bm{\theta}\in\Omega(\mathcal{C})}
J_{\mathrm{phys}}
\left(
\mathcal{D}(G;\mathcal{C},\bm{\theta})
\right).
\end{equation}
Here, $J_{\mathrm{phys}}(\cdot)$ denotes a high-fidelity physics-based performance objective evaluated by a physics solver. 
In the automotive aerodynamic instantiation, the physics solver is a high-fidelity CFD solver and $J_{\mathrm{phys}}$ is the drag coefficient $C_d$. 
Since evaluating $J_{\mathrm{phys}}$ for every candidate design is computationally expensive, we solve the problem using a knowledge-constrained, physics-solver-in-the-loop, surrogate-assisted optimization workflow.
The preservation constraints $\mathcal{H}$ are not imposed as separate analytic inequalities in the present implementation.
Instead, they are embedded in the construction of the editable control boxes $\mathcal{C}$, the admissible deformation space $\Omega(\mathcal{C})$, and the deformation operator $\mathcal{D}$.

\subsection{Framework overview}
\label{subsec:framework_overview}

\begin{figure}
    \centering
    \includegraphics[width=\textwidth]{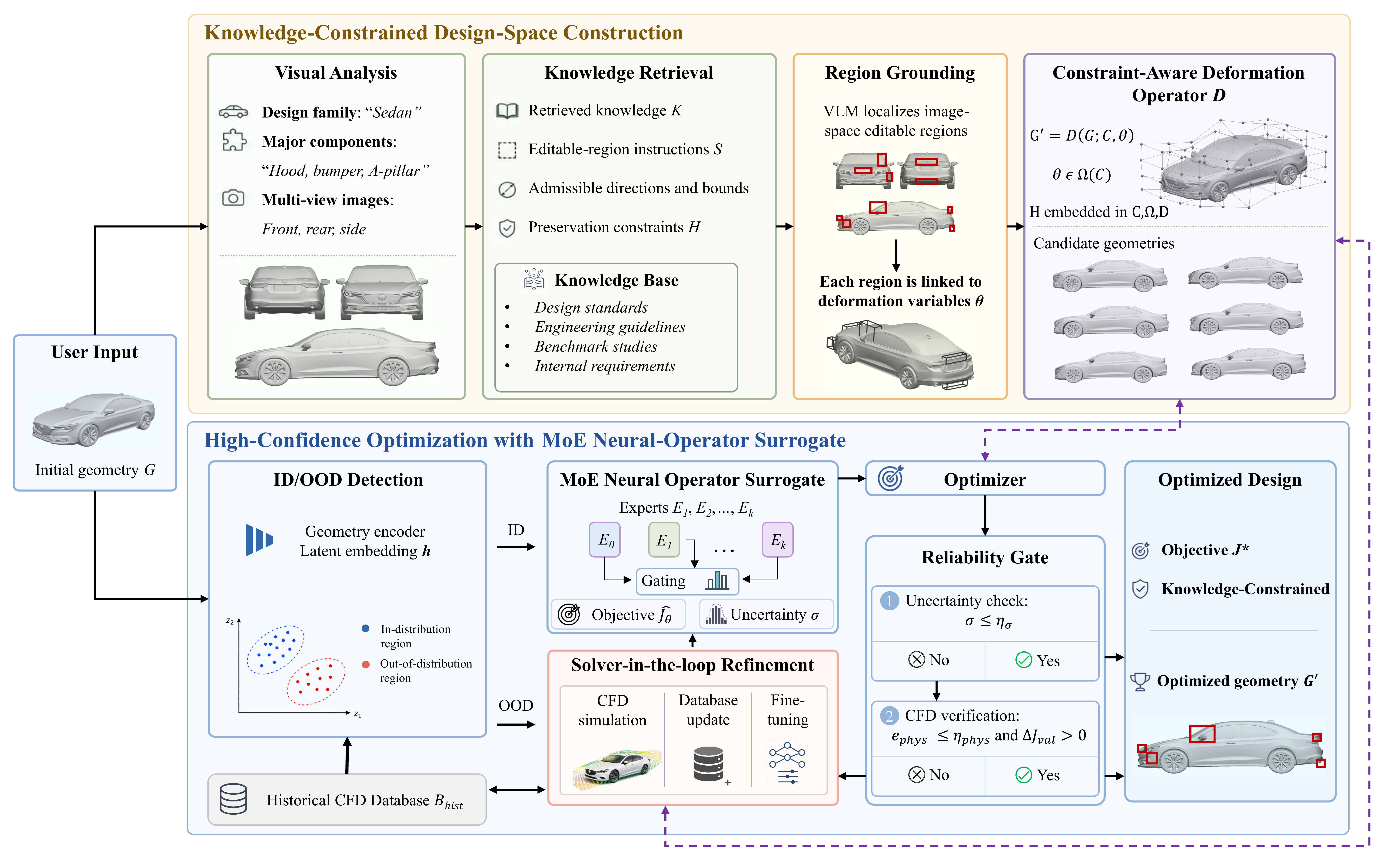}
    \caption{Overview of the framework.}
    \label{fig:framework}
\end{figure}

The proposed framework consists of two tightly coupled modules, as illustrated in Fig.~\ref{fig:framework}.

The first module constructs a knowledge-constrained design space for the initial geometry $G$.
It starts with the VLM analyzing the multi-view images $\mathcal{I}$ of $G$ to identify the design family and major components; the RAG module then retrieves the relevant domain knowledge $\mathcal{K}$, which the LLM translates into design instructions, including editable components, allowable deformation directions, deformation bounds, and preservation constraints.
Then, given the multi-view images $\mathcal{I}$ of $G$ and the LLM-generated instructions, the VLM localizes the semantic editable regions in the rendered views; these view-wise regions are then fused into the three-dimensional editable control-box set $\mathcal{C}$ and associated with deformation variables $\bm{\theta}$.
These control boxes and variables are further used to define a constraint-aware deformation operator $\mathcal{D}$.
Together with the retrieved design knowledge, this process specifies the admissible deformation space $\Omega(\mathcal{C})$ and the preservation constraints $\mathcal{H}$.
The resulting outputs, namely $\mathcal{D}$, $\Omega(\mathcal{C})$, and $\mathcal{H}$, are then passed to the downstream optimization module.

The second module integrates OOD detection, solver-based labeling, MoE-NO surrogate prediction, and Mahalanobis-percentile uncertainty estimation (Section~\ref{sec:online}) into a closed-loop optimization process.
Importantly, the physics solver is not invoked for every candidate design.
Instead, solver calls that feed back into the optimization are triggered only through two decision gates.

The first gate operates at the distribution level.
The encoder of the MoE-NO surrogate maps each query geometry into a latent representation and checks whether it is sufficiently covered by the historical simulation database.
If the query geometry is detected as out-of-distribution, the deformation operator $\mathcal{D}$ is activated to generate local samples around the query geometry within the admissible deformation space $\Omega(\mathcal{C})$, slightly expanded for surrogate enrichment.
These samples are evaluated by the physics solver and used to adapt the surrogate to the newly discovered design family.
The MoE-NO surrogate is then fine-tuned with the newly generated solver-labeled samples, and a new expert branch is introduced to represent the newly explored design region.

The second gate operates at the optimization level.
During optimization, the MoE-NO surrogate predicts the objective value of each candidate geometry.
For the surrogate optimum, its reliability is quantified by a Mahalanobis-percentile uncertainty score computed in the frozen latent space.
If the uncertainty score is below a prescribed threshold, the candidate is considered sufficiently supported by the current training database and can be accepted without an additional solver call during the optimization loop.
Otherwise, the candidate is sent to the high-fidelity physics solver for verification.
If the solver result agrees with the surrogate prediction within a prescribed relative-error tolerance and confirms an improvement over the baseline, the candidate is accepted.
If not, additional solver-labeled samples are generated in a local neighborhood of the candidate for surrogate enrichment and added to the training database. The MoE-NO surrogate is fine-tuned using the enriched database, and the optimization is restarted over the original admissible deformation space $\Omega(\mathcal{C})$.

The implementation details are presented in the following sections.

\subsection{Knowledge-constrained design-space construction}

This section describes how the first module in Fig.~\ref{fig:framework} converts domain knowledge into semantic editable regions, fused three-dimensional editable control boxes, admissible deformation bounds, and constraint-aware deformation variables.

\subsubsection{Retrieval and synthesis of knowledge}

The overall workflow for grounding the semantic editable regions, the fused three-dimensional editable control-box set $\mathcal{C}$,  the admissible deformation space $\Omega(\mathcal{C})$, and the preservation constraints $\mathcal{H}$ is illustrated in Fig.~\ref{fig:framework}. 
Given the multi-view images $\mathcal{I}=\{I_v\}_{v=1}^{N_v}$, the VLM first analyzes the visual inputs and identifies the design family and major components in natural-language form~\cite{radford2021learning}. 
Then, a RAG module~\cite{lewis2020retrieval,gao2023retrieval} retrieves relevant domain knowledge, including design standards, engineering guidelines, benchmark studies, and internal design requirements. 
The retrieved knowledge is synthesized into three types of outputs: a set of textual standards for the semantic editable regions,
\begin{equation}
    \mathcal{S}
    =
    \left\{
    s_m
    \right\}_{m=1}^{N_c},
\end{equation}
a set of local admissible deformation ranges,
\begin{equation}
   \left\{
    \Omega_m(c_m)
    \right\}_{m=1}^{N_c},
\end{equation}
and a set of preservation constraints,
\begin{equation}
    \mathcal{H}
    =
    \left\{
    \chi_q
    \right\}_{q=1}^{N_H}.
\end{equation}
Here, each textual standard $s_m$ describes the semantic meaning, geometric scope, admissible deformation direction, and design requirements of the $m$-th editable region.
Each $\Omega_m(c_m)$ specifies the corresponding lower and upper deformation bounds after the semantic region has been converted into the physical control box $c_m$, while $\chi_q$ collects the non-editable or weakly editable constraints that preserve packaging, safety, manufacturability, component interfaces, smoothness, and design identity.
These knowledge-derived standards guide the VLM to localize each editable region in every view by predicting an axis-aligned bounding rectangle in the pixel coordinate system:
\begin{equation}
    \mathcal{C}^{p}_{v}
    =
    \left\{
    \left(
    \bm{x}_{m,v}^{\min,p},
    \bm{x}_{m,v}^{\max,p}
    \right)
    \right\}_{m=1}^{N_c},
    \qquad
    v=1,2,\cdots,N_v .
    \label{eq:ctrl_image}
\end{equation}
The subscript $v$ denotes the $v$-th image view, while the superscript $p$ indicates pixel coordinates. 
The vectors $\bm{x}_{m,v}^{\min,p}$ and $\bm{x}_{m,v}^{\max,p}$ represent the minimum and maximum pixel coordinates of the predicted bounding rectangle, respectively.

In this work, the rendered views are generated from calibrated orthographic camera settings.
Therefore, each image-space bounding rectangle can be mapped to the corresponding physical coordinate interval on the two axes visible in that view.
The remaining coordinate direction is recovered by combining annotations from other orthographic views.
When the same region is observed in multiple views, the view-wise physical intervals are fused into a conservative 3D enclosing box.
This procedure does not infer a unique 3D point from a single 2D pixel location; instead, it uses calibrated multi-view projections to recover a compact physical control box.
To express the control boxes in physical coordinates, a view-specific linear transformation $\mathcal{L}_v$ is applied to each image-space rectangle:
\begin{equation}
    \mathcal{C}_{v}
    =
    \mathcal{L}_v
    \left(
    \mathcal{C}^{p}_{v}
    \right)
    =
    \left\{
    \left(
    \mathcal{L}_v
    \left(
    \bm{x}_{m,v}^{\min,p}
    \right),
    \mathcal{L}_v
    \left(
    \bm{x}_{m,v}^{\max,p}
    \right)
    \right)
    \right\}_{m=1}^{N_c}.
\end{equation}
Therefore,
\begin{equation}
    \mathcal{C}_{v}
    =
    \left\{
    \left(
    \bm{x}_{m,v}^{\min},
    \bm{x}_{m,v}^{\max}
    \right)
    \right\}_{m=1}^{N_c},
    \qquad
    v=1,2,\cdots,N_v ,
\end{equation}
where $\bm{x}_{m,v}^{\min}$ and $\bm{x}_{m,v}^{\max}$ are the physical-coordinate bounds of the $m$-th view-wise control-region proposal observed from the $v$-th view.

Since the same editable region may be visible in multiple views, the final physical control box is obtained by aggregating the view-wise bounds. 
Specifically, for each control box, we compute the component-wise minimum and maximum over all views:
\begin{align}
    \bm{x}_{m}^{\min}
    &=
    \operatorname*{cmin}_{v=1}^{N_v}
    \bm{x}_{m,v}^{\min},
    \\
    \bm{x}_{m}^{\max}
    &=
    \operatorname*{cmax}_{v=1}^{N_v}
    \bm{x}_{m,v}^{\max},
\end{align}
where $\operatorname*{cmin}$ and $\operatorname*{cmax}$ denote component-wise minimum and maximum operations, respectively. 
The final editable control boxes are then parameterized by their center locations and side lengths:
\begin{equation}
    \mathcal{C}
    =
    \left\{
    c_m
    \right\}_{m=1}^{N_c},
\end{equation}
with
\begin{align}
    \bm{x}_{m}^{c}
    &=
    \frac{
    \bm{x}_{m}^{\max}
    +
    \bm{x}_{m}^{\min}
    }{2},
    \\
    \bm{l}_{m}
    &=
    \bm{x}_{m}^{\max}
    -
    \bm{x}_{m}^{\min}.
\end{align}
Here, $c_m=(\bm{x}_{m}^{c},\bm{l}_{m})$, $\bm{x}_{m}^{c}$ denotes the center of the $m$-th editable control box, and $\bm{l}_{m}$ denotes its physical size.

In this work, we use \texttt{Qwen 3.5-Plus} as the LLM and \texttt{Gemini 2.5 Flash Image} as the VLM. 
For the automotive case study, the RAG knowledge base includes knowledge related to safety~\cite{crandall2002designing,hu2024association}, aesthetics~\cite{feldinger2017automotive,qi2024evaluation}, manufacturability~\cite{pinfold2001application,naiju2021dfma}, and aerodynamic drag~\cite{thomas2016adjoint,sudin2014review}. 
The public portion of the knowledge base is constructed from road-vehicle aerodynamic studies, benchmark-model literature, and automotive aerodynamic textbooks~\cite{ahmed1984some,heft2012introduction,ekman2020assessment,hucho2013aerodynamics,schuetz2015aerodynamics,barnard2001road}. 
The benchmark sources include the Ahmed body and DrivAer model literature~\cite{ahmed1984some,heft2012introduction,collin2016numerical}. 
The non-public portion consists of proprietary company documents that provide additional industrial design knowledge; these documents are not released due to confidentiality restrictions.

\subsubsection{Constraint-aware geometric deformation}

Here we show the details of the implementation of the constraint-aware deformation operator $\mathcal{D}$ under the knowledge-derived constraints $(\mathcal{C},\Omega(\mathcal{C}),\mathcal{H})$.

To ensure a smooth and continuous deformation process and explicitly constrain the local deformation range, the DFFD method~\cite{sederberg1986free,hsu1992direct,lamousin1994nurbs} is used in this work.
The key idea of DFFD is to represent the deformation as a smooth continuous spatial displacement field.
Instead of independently moving each selected node, DFFD infers a globally smooth deformation field $\bm{d}$ from the prescribed local displacement constraints $\bm{\theta}$, so that nearby nodes deform coherently and the local deformation range can be explicitly bounded.
Because DFFD represents deformation as a smooth spatial displacement field, nodes outside the editable regions may also receive small induced displacements.
Therefore, the method is constraint-aware in the sense that the prescribed design variables are applied only through the knowledge-grounded editable control boxes $\mathcal{C}$ and bounded admissible deformation space $\Omega(\mathcal{C})$, while the constraints $\mathcal{H}$ are enforced through bounded local variables, unchanged topology, and smoothness regularization.
It does not impose an exact zero-displacement constraint on every non-editable node.

\begin{figure}
    \centering
    \includegraphics[width=0.95\textwidth]{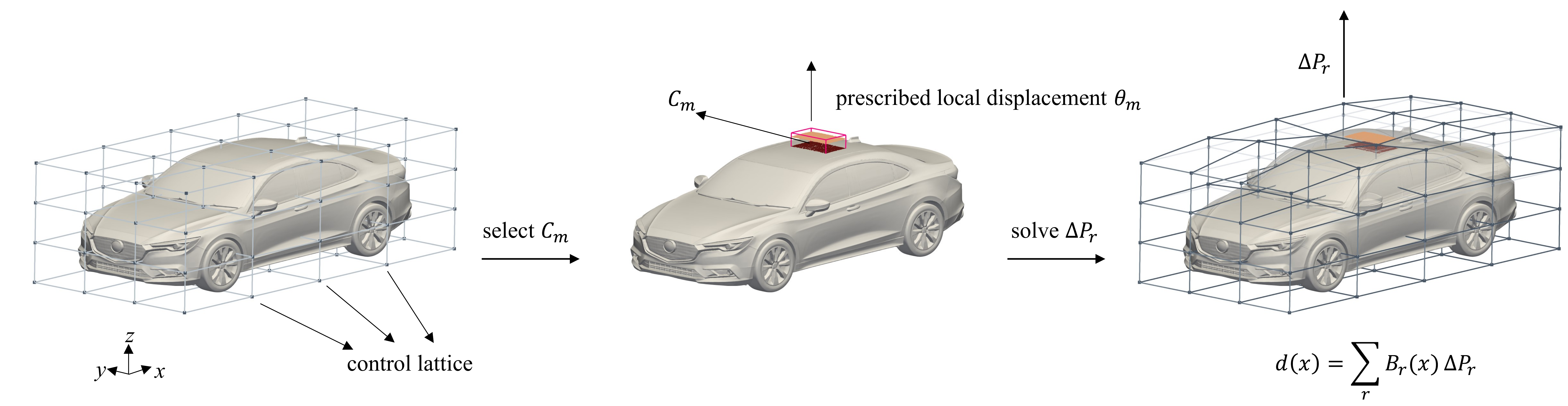}
    \caption{DFFD-based constrained geometric deformation.}
    \label{fig:dffd}
\end{figure}

The details of the DFFD method are illustrated in Fig.~\ref{fig:dffd}.
For each editable control-box parameter $c_m \in \mathcal{C}$, we define the corresponding axis-aligned physical region as
\begin{equation}
    \mathcal{R}_m =
    \left\{
    \bm{x}\in\mathbb{R}^3
    \mid
    \bm{x}_m^c-\frac{\bm{l}_m}{2}
    \preceq
    \bm{x}
    \preceq
    \bm{x}_m^c+\frac{\bm{l}_m}{2}
    \right\}.
\end{equation}

Considering the geometry represented by Eq.~\eqref{eq:geo}, DFFD only modifies
the coordinates of the nodes $\bm{n}_k$ while keeping the edge and face
connectivity unchanged. 
The constrained nodes inside $\mathcal{R}_m$ are selected as
\begin{equation}
    \mathcal{N}_m =
    \left\{
    \bm{n}_k
    \mid
    \bm{n}_k \in \mathcal{N},\;
    \bm{n}_k \in \mathcal{R}_m
    \right\}.
\end{equation}

To parameterize the continuous displacement field, DFFD first embeds the original geometry $G$ into a regular control lattice defined over a spatial domain $\mathcal{R}_l$, as shown in Fig.~\ref{fig:dffd}. 
In this work, $\mathcal{R}_l$ is chosen as the bounding box that encloses the original node set $\mathcal{N}$:
\begin{equation}
    \mathcal{R}_l
    =
    \left\{
    \bm{x}\in\mathbb{R}^3
    \mid
    \bm{x}^{\min}
    \preceq
    \bm{x}
    \preceq
    \bm{x}^{\max}
    \right\},
\end{equation}
where $\bm{x}^{\min}$ and $\bm{x}^{\max}$ denote the component-wise minimum and
maximum coordinates of all nodes in $\mathcal{N}$, respectively. 
A set of control points $\{\bm{P}_r\}_{r=1}^{R}$ is then placed in $\mathcal{R}_l$ to define the deformation field, where $R$ is the number of control points.

Let $\Delta \bm{P}_r$ denote the
displacement of the $r$-th control point, and let $B_r(\bm{n}_k)$ be the
corresponding basis weight evaluated at node $\bm{n}_k$. The smooth
displacement field for each node $\bm{n}_k$ is expressed as
\begin{equation}
    \bm{d}(\bm{n}_k)
    =
    \sum_{r=1}^{R}
    B_r(\bm{n}_k)\Delta \bm{P}_r.
\end{equation}

Then, we require the displacement field of the constrained nodes in $\mathcal{N}_m$ to be close to the prescribed local displacement vector $\bm{\theta}_m$.
Therefore, the control point displacements are obtained by solving
\begin{equation}
    \min_{\{\Delta \bm{P}_r\}_{r=1}^{R}}
    \sum_{m=1}^{N_c}
    \sum_{\bm{n}_k\in\mathcal{N}_m}
    \left\|
    \sum_{r=1}^{R}
    B_r(\bm{n}_k)\Delta \bm{P}_r
    -
    \bm{\theta}_m
    \right\|_2^2
    +
    \beta
    \left\|
    \bm{L}\Delta \bm{p}
    \right\|_2^2,
\end{equation}
where $\Delta \bm{p}=[\Delta \bm{P}_1^\top,\Delta \bm{P}_2^\top,\dots,\Delta \bm{P}_R^\top]^\top$ stacks all control point displacements, $\bm{L}$ is a smoothness regularization operator, and $\beta$ controls the strength of the smoothness constraint.

Finally, the deformed geometry is obtained by applying the solved displacement field to all nodes:
\begin{equation}
    \bm{n}_k'
    =
    \bm{n}_k + \bm{d}(\bm{n}_k).
\end{equation}
The output geometry is then written as
\begin{equation}
G'(\bm{\theta})
=
\left(
\mathcal{F},
\mathcal{E},
\mathcal{N}'
\right),
\qquad
\mathcal{N}'
=
\left\{
\bm{n}'_k
\right\}_{k=1}^{N_n}.
\end{equation}
The face and edge connectivity remain unchanged, and only the node coordinates are updated.

\subsection{High-confidence Optimization with the MoE-NO Surrogate}

\subsubsection{MoE-NO surrogate}
\label{sec:moe}

\begin{figure}
    \centering
    \includegraphics[width=0.95\textwidth]{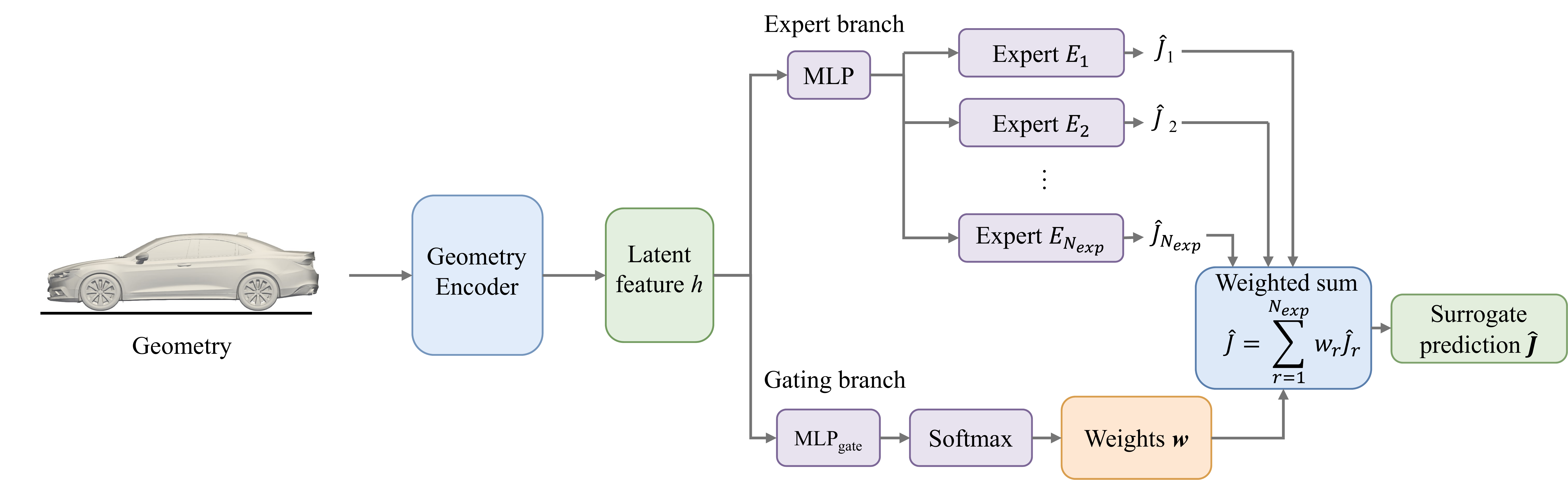}
    \caption{MoE-NO surrogate for performance prediction and uncertainty estimation.}
    \label{fig:moe}
\end{figure}

The encoder-based MoE-NO surrogate is illustrated in Fig.~\ref{fig:moe}.
The surrogate model maps the input geometry to the target performance quantity.
The input to the neural operator is the geometry $G$. 
For encoding, $G$ is represented by its surface node set $\mathcal{N}=\{\bm{n}_k\}_{k=1}^{N_n}$ in Eq.~\eqref{eq:geo}, i.e., 
the surface point cloud, whereas the face and edge connectivity $\mathcal{F}$ and $\mathcal{E}$ are not used as encoder input.
For notational simplicity, this point-cloud input is still denoted by $G$. A geometry encoder first
extracts a compact latent representation,
\begin{equation}
\bm{h}
=
\mathrm{Encoder}
\left(
G
\right),
\qquad
\bm{h}\in\mathbb{R}^{N_h},
\label{eq:encoder}
\end{equation}
where $N_h$ denotes the dimension of the latent space. In this work, the encoder is inherited from the encoder module of our previously proposed neural operator, DragSolver~\cite{liu2025dragsolver}.
After the encoder, the latent feature $\bm{h}$ is passed into two branches.
The first branch consists of $N_{\mathrm{exp}}$ expert networks. Each expert
produces a candidate prediction based on the latent feature:
\begin{equation}
\hat{J}_{\phi,r}^{\mathrm{exp}}
=
E_r
\left(
\mathrm{MLP}_{\mathrm{exp}}
\left(
\bm{h}
\right)
\right),
\qquad
r=1,\ldots,N_{\mathrm{exp}} .
\end{equation}
Here, $\hat{J}_{\phi,r}^{\mathrm{exp}}$ is the performance prediction produced by the $r$-th expert, and $E_r$ denotes the corresponding expert network, and $\mathrm{MLP}_{\mathrm{exp}}$ denotes a multi-layer perceptron (MLP) applied to the latent feature.The subscript $\phi$ collects the trainable parameters of the MoE-NO surrogate, including the encoder, the experts, and the gating network.
The second branch is a gating network that assigns an adaptive routing weight to each expert:
\begin{equation}
\bm{w}
=
\mathrm{Softmax}
\left(
\mathrm{MLP}_{\mathrm{gate}}
\left(
\bm{h}
\right)
\right),
\qquad
\bm{w}\in\mathbb{R}^{N_{\mathrm{exp}}},
\end{equation}
where $\bm{w}=(w_1,\ldots,w_{N_{\mathrm{exp}}})$ is the expert-weight vector and satisfies
\begin{equation}
w_r\geq 0,
\qquad
\sum_{r=1}^{N_{\mathrm{exp}}}w_r=1.
\end{equation}
The final surrogate prediction is obtained as the weighted combination of all
expert predictions:
\begin{equation}
\hat{J}_{\phi}
=
\sum_{r=1}^{N_{\mathrm{exp}}}
w_r
\hat{J}_{\phi,r}^{\mathrm{exp}} .
\label{eq:moe_prediction}
\end{equation}
In the aerodynamic case considered here, $\hat{J}_{\phi}$ predicts the drag coefficient $C_d$.

\subsubsection{OOD detection}
\label{sec:detection}

Once the MoE-NO surrogate has been trained on the existing historical simulation database, the encoder defined in Eq.~\eqref{eq:encoder} is used to perform OOD detection.
Specifically, for a new query geometry $G_q$ and the historical database $\mathcal{B}_{\mathrm{hist}}$, the query geometry and all historical geometries are encoded into the latent space as
\begin{align}
    \bm{h}_q &= \mathrm{Encoder}(G_q),\\
    \bm{h}_i &= \mathrm{Encoder}(G^{(i)}),
    \qquad
    i = 1,2,\cdots,N_d,
\end{align}
where $N_d$ is the total number of samples in the historical database and $G_q$ denotes the query geometry for distribution-level assessment.

Three complementary criteria are used for comprehensive OOD detection: including the nearest-neighbor Euclidean distance, the $N_{\mathrm{KNN}}$-nearest-neighbor mean distance (KNN), and the maximum cosine similarity \cite{kaya2019deep,pimentel2014review}.
The Euclidean distance between the query embedding $\bm{h}_q$ and the $i$-th historical embedding $\bm{h}_i$ is defined as
\begin{equation}
d_{i,\mathrm{Euc}}
=
\left\|
\bm{h}_q
-
\bm{h}_i
\right\|_2 .
\end{equation}
The nearest-neighbor Euclidean score is then computed as
\begin{equation}
d_{\mathrm{Euc}}
=
\min_{i=1}^{N_d}
d_{i,\mathrm{Euc}} .
\end{equation}

To further evaluate the local database coverage around the query geometry, the local-neighborhood score is computed as the mean distance to the $N_{\mathrm{KNN}}$ nearest historical embeddings:
\begin{equation}
d_{\mathrm{KNN}}
=
\frac{1}{N_{\mathrm{KNN}}}
\sum_{i}
\left\|
\bm{h}_q - \bm{h}_i
\right\|_2,
\end{equation}
where the sum runs over the $N_{\mathrm{KNN}}$ nearest historical embeddings of $\bm{h}_q$.
In this work, $N_{\mathrm{KNN}} = 5$ is treated as a hyperparameter.

The maximum cosine similarity measures the angular similarity between the query geometry and the historical geometries:
\begin{equation}
s_{\mathrm{cos}}
=
\max_{i=1}^{N_d}
\frac{
\bm{h}_i^{\top}\bm{h}_q
}{
\|\bm{h}_i\|_2
\|\bm{h}_q\|_2
+
\varepsilon
},
\end{equation}
where $\varepsilon>0$ is a small constant introduced to avoid division by zero.

The decision thresholds are calibrated from the historical latent embeddings using a leave-one-out percentile strategy~\cite{chandola2009anomaly,pimentel2014review}.
This calibration estimates the typical distance and similarity ranges among in-distribution historical samples, so that the OOD thresholds are determined from the database itself rather than selected manually.
For the $i$-th historical embedding, $\bm{h}_i$ is treated as a pseudo-query and the remaining historical embeddings, $\{\bm{h}_j\}_{j\neq i}$, are used as the reference set.
The resulting leave-one-out Euclidean, KNN, and cosine calibration scores are denoted by
$d_{\mathrm{Euc}}^{\mathrm{LOO}}(i)$,
$d_{\mathrm{KNN}}^{\mathrm{LOO}}(i)$, and
$s_{\mathrm{cos}}^{\mathrm{LOO}}(i)$, respectively.
Collecting these scores over all historical embeddings gives
\begin{align}
\mathcal{D}_{\mathrm{Euc}}^{\mathrm{LOO}}
&=
\left\{
d_{\mathrm{Euc}}^{\mathrm{LOO}}(i)
\right\}_{i=1}^{N_d},
\\
\mathcal{D}_{\mathrm{KNN}}^{\mathrm{LOO}}
&=
\left\{
d_{\mathrm{KNN}}^{\mathrm{LOO}}(i)
\right\}_{i=1}^{N_d},
\\
\mathcal{S}_{\mathrm{cos}}^{\mathrm{LOO}}
&=
\left\{
s_{\mathrm{cos}}^{\mathrm{LOO}}(i)
\right\}_{i=1}^{N_d}.
\end{align}
Given a calibration percentile $p_{\mathrm{cal}}\in(0,100)$, the thresholds are defined as
\begin{align}
\tau_{\mathrm{Euc}}
&=
Q_{p_{\mathrm{cal}}}
\left(
\mathcal{D}_{\mathrm{Euc}}^{\mathrm{LOO}}
\right),
\\
\tau_{\mathrm{KNN}}
&=
Q_{p_{\mathrm{cal}}}
\left(
\mathcal{D}_{\mathrm{KNN}}^{\mathrm{LOO}}
\right),
\\
\tau_{\mathrm{cos}}
&=
Q_{100-p_{\mathrm{cal}}}
\left(
\mathcal{S}_{\mathrm{cos}}^{\mathrm{LOO}}
\right),
\end{align}
where $Q_{p_{\mathrm{cal}}}(\cdot)$ denotes the $p_{\mathrm{cal}}$-th percentile.

For the query geometry $G_q$, three binary indicators are evaluated as
\begin{align}
I_{\mathrm{Euc}}
&=
\mathbb{I}
\left[
d_{\mathrm{Euc}}
\leq
\tau_{\mathrm{Euc}}
\right],
\\
I_{\mathrm{KNN}}
&=
\mathbb{I}
\left[
d_{\mathrm{KNN}}
\leq
\tau_{\mathrm{KNN}}
\right],
\\
I_{\mathrm{cos}}
&=
\mathbb{I}
\left[
s_{\mathrm{cos}}
\geq
\tau_{\mathrm{cos}}
\right].
\end{align}

Here, $\mathbb{I}[\cdot]$ is the indicator function, which equals one if the bracketed condition holds and zero otherwise.The final decision is made using a two-out-of-three voting rule \cite{zhou2012ensemble},
\begin{equation}
I_{\mathrm{hist}}
=
\mathbb{I}
\left[
I_{\mathrm{Euc}}
+
I_{\mathrm{KNN}}
+
I_{\mathrm{cos}}
\geq
2
\right].
\end{equation}
The query geometry is classified as in-distribution (ID) when $I_{\mathrm{hist}}=1$ and OOD otherwise.

\subsubsection{OOD-aware resampling and model fine-tuning}

When a new query geometry $G_q$ is detected as OOD, direct reuse of the existing surrogate may no longer provide reliable predictions.
This condition activates the first physics-solver-in-the-loop mechanism.
To adapt the surrogate to the newly encountered design family, we construct a local high-fidelity dataset around $G_q$ and use it to fine-tune the MoE-NO surrogate.

Specifically, the deformation operator $\mathcal{D}$ is used to generate $N_s$ locally deformed geometries from $G_q$.

To improve the robustness of the surrogate near the boundary of the admissible region, the sampling range used for surrogate-data enrichment is slightly expanded beyond the original deformation bounds.
This expanded range is used only for local surrogate adaptation and does not modify theadmissible optimization space $\Omega(\mathcal{C})$.
For each editable control box $c_m$, the expanded sampling space is defined as
\begin{equation}
\Omega_m^{\mathrm{ex}}(c_m)
=
\left\{
\bm{\theta}_m\in\mathbb{R}^3
\mid
\bm{\theta}_m^l
-
\alpha_{\mathrm{ex}}
\left(
\bm{\theta}_m^u-\bm{\theta}_m^l
\right)
\preceq
\bm{\theta}_m
\preceq
\bm{\theta}_m^u
+
\alpha_{\mathrm{ex}}
\left(
\bm{\theta}_m^u-\bm{\theta}_m^l
\right)
\right\},
\end{equation}
where $\alpha_{\mathrm{ex}}=0.1$ is the expansion factor.

The complete sampling domain for OOD adaptation is then
\begin{equation}
\Omega_{\mathrm{sam}}^{\mathrm{ex}}(\mathcal{C})
=
\prod_{m=1}^{N_c}
\Omega_m^{\mathrm{ex}}(c_m).
\end{equation}

Latin hypercube sampling (LHS)~\cite{mckay2000comparison,morris1995exploratory} is adopted to sample the local deformation vectors
\begin{equation}
\bm{\theta}_{\mathrm{new}}^{(i)}
\in
\Omega_{\mathrm{sam}}^{\mathrm{ex}}(\mathcal{C}),
\qquad
i=1,\ldots,N_s .
\end{equation}
For the $i$-th sampled deformation parameter $\bm{\theta}_{\mathrm{new}}^{(i)}$, the corresponding geometry is generated as
\begin{equation}
G_{\mathrm{new}}^{(i)}
=
\mathcal{D}
\left(
G_q;\mathcal{C},\bm{\theta}_{\mathrm{new}}^{(i)}
\right),
\qquad
i=1,\ldots,N_s .
\end{equation}
The objective performance of each generated geometry is evaluated using the high-fidelity physics solver.
The newly sampled high-fidelity dataset is therefore given by
\begin{equation}
\mathcal{B}_{\mathrm{new}}
=
\left\{
\left(
G_{\mathrm{new}}^{(i)},
J_{\mathrm{phys}}
\left(
G_{\mathrm{new}}^{(i)}
\right)
\right)
\right\}_{i=1}^{N_s}.
\end{equation}

To represent the response of the newly encountered design family, a new expert branch $E_{N_{\mathrm{exp}}+1}$ is added to the original MoE-NO surrogate.
The expanded expert set becomes
\begin{equation}
\mathcal{E}_{\mathrm{exp}}^{+}
=
\left\{
E_1,
E_2,
\ldots,
E_{N_{\mathrm{exp}}},
E_{N_{\mathrm{exp}}+1}
\right\}.
\end{equation}
The adaptation training database is then updated as
\begin{equation}
\mathcal{B}_{\mathrm{adapt}}
=
\mathcal{B}_{\mathrm{hist}}
\cup
\mathcal{B}_{\mathrm{new}}^{\mathrm{train}},
\end{equation}
where $\mathcal{B}_{\mathrm{new}}^{\mathrm{train}}\subseteq\mathcal{B}_{\mathrm{new}}$ denotes the OOD adaptation subset.
During fine-tuning, the newly added expert $E_{N_{\mathrm{exp}}+1}$ is trained to capture the local geometry--performance relationship around $G_q$, while the previously learned experts retain the knowledge from the historical simulation database.
The gate network is also updated so that samples from the newly discovered design family can be properly routed to the new expert branch.

\subsubsection{Uncertainty-gated online refinement}
\label{sec:online}
This study uses an uncertainty-gated online refinement loop to prevent the surrogate optimizer from accepting designs that are outside the reliable region of the current training database.
Surrogate-assisted evolutionary optimization is commonly used when direct evaluations are expensive~\cite{liang2025survey}.
Here, the MoE-NO surrogate provides an approximation of the physics objective defined in Section~\ref{subsec:general_problem_formulation}, while a Mahalanobis-percentile score decides whether the predicted optimum should be accepted directly or checked by the high-fidelity solver.

Let $f_{\phi}$ denote the current MoE-NO surrogate.
For a deformation-variable set $\bm{\theta}\in\Omega(\mathcal{C})$, the generated geometry and surrogate approximation are written as:
\begin{equation}
G'(\bm{\theta})
=
\mathcal{D}
\left(
G;\mathcal{C},\bm{\theta}
\right),
\qquad
\hat{J}_{\phi}(\bm{\theta})
=
f_{\phi}
\left(
G'(\bm{\theta})
\right)
\approx
J_{\mathrm{phys}}
\left(
G'(\bm{\theta})
\right).
\end{equation}
In the aerodynamic case considered here, $\hat{J}_{\phi}$ predicts the drag coefficient $C_d$.

The reliability of $G'(\bm{\theta})$ is evaluated in the frozen encoder latent space.
Let $\mathcal{B}_{\mathrm{cur}}$ denote the current training database available to the surrogate.
For ID optimization, $\mathcal{B}_{\mathrm{cur}}=\mathcal{B}_{\mathrm{hist}}$.
After OOD adaptation or online refinement, $\mathcal{B}_{\mathrm{cur}}$ is replaced by the corresponding enriched database.
The reference embeddings extracted from $\mathcal{B}_{\mathrm{cur}}$ are denoted by
\begin{equation}
\mathcal{H}_{\mathrm{ref}}
=
\left\{
\tilde{\bm{h}}_i^{\mathrm{ref}}
\right\}_{i=1}^{N_{\mathrm{ref}}},
\end{equation}
where $N_{\mathrm{ref}}$ is the number of reference embeddings in the current training database, and $\tilde{\bm{h}}_i^{\mathrm{ref}}$ denotes the standardized latent embedding of the $i$-th reference geometry.

Following Mahalanobis-distance-based confidence estimation~\cite{lee2018simple}, the latent vector of the candidate is compared with the reference latent vectors extracted from the currently available training database.
Let $\tilde{\bm{h}}(\bm{\theta})$ denote the standardized latent embedding of $G'(\bm{\theta})$, and let $\bm{\mu}$ and $\bm{\Sigma}$ denote the mean and Ledoit--Wolf-shrunk covariance matrix~\cite{ledoit2004well} estimated from the standardized reference embeddings.
The Mahalanobis distance of a standardized latent vector $\tilde{\bm{h}}$ is defined as
\begin{equation}
d_M(\tilde{\bm{h}})
=
\sqrt{
\left(
\tilde{\bm{h}}-\bm{\mu}
\right)^\top
\bm{\Sigma}^{-1}
\left(
\tilde{\bm{h}}-\bm{\mu}
\right)
}.
\end{equation}
For a candidate geometry $G'(\bm{\theta})$, this gives
\begin{equation}
d_M(\bm{\theta})
=
d_M
\left(
\tilde{\bm{h}}(\bm{\theta})
\right).
\end{equation}
For the $i$-th reference embedding, the corresponding reference distance is
\begin{equation}
d_{M,i}^{\mathrm{ref}}
=
d_M
\left(
\tilde{\bm{h}}_i^{\mathrm{ref}}
\right).
\end{equation}
The candidate distance is converted into the empirical percentile score
\begin{equation}
\sigma
\left(
\bm{\theta}
\right)
=
\frac{1}{N_{\mathrm{ref}}}
\sum_{i=1}^{N_{\mathrm{ref}}}
\mathbb{I}
\left[
d_{M,i}^{\mathrm{ref}}
\leq
d_M
\left(
\bm{\theta}
\right)
\right],
\qquad
\sigma(\bm{\theta})\in[0,1].
\label{eq:mahalanobis_percentile}
\end{equation}
A small $\sigma$ means that the candidate is well covered by the available data, whereas a large $\sigma$ indicates a tail-region design that requires solver verification.

The optimizer first searches the same knowledge-constrained design space as the original problem, but replaces the expensive physics objective with its surrogate approximation:
\begin{equation}
\bm{\theta}_{\mathrm{sur}}^{*}
=
\arg\min_{\bm{\theta}\in\Omega(\mathcal{C})}
\hat{J}_{\phi}
\left(
\bm{\theta}
\right).
\label{eq:surrogate_optimum}
\end{equation}
In principle, the optimizer can be instantiated by standard derivative-free optimizers, such as differential evolution (DE), genetic algorithms (GA), or Bayesian optimization (BO)~\cite{storn1997differential,holland1992adaptation,jones1998efficient,shahriari2015taking}. 
In the numerical experiments, we use DE unless otherwise stated.
If $\sigma(\bm{\theta}_{\mathrm{sur}}^{*})\leq\eta_{\sigma}$, the candidate is considered sufficiently supported by the current training database and no new physics solve is required inside the optimization loop.
In this work, $\eta_{\sigma}=0.6$, and the accepted design is \begin{equation}
G^{*}
=
G'
\left(
\bm{\theta}_{\mathrm{sur}}^{*}
\right).
\end{equation}

If the uncertainty exceeds the threshold, the candidate is evaluated by the high-fidelity solver:
\begin{equation}
J_{\mathrm{val}}
=
J_{\mathrm{phys}}
\left(
G'
\left(
\bm{\theta}_{\mathrm{sur}}^{*}
\right)
\right).
\end{equation}
The relative surrogate--physics discrepancy is then measured as:
\begin{equation}
e_{\mathrm{phys}}
=
\frac{
\left|
\hat{J}_{\phi}
\left(
\bm{\theta}_{\mathrm{sur}}^{*}
\right)
-
J_{\mathrm{val}}
\right|
}{
\left|
J_{\mathrm{val}}
\right|
+
\varepsilon
}.
\end{equation}
The validated improvement over the baseline geometry is measured by
\begin{equation}
\Delta J_{\mathrm{val}}
=
\frac{
J_{\mathrm{phys}}(G)-J_{\mathrm{val}}
}{
J_{\mathrm{phys}}(G)
}.
\end{equation}
If $e_{\mathrm{phys}}\leq\eta_{\mathrm{phys}}$ and $\Delta J_{\mathrm{val}}>0$, the candidate is accepted as a solver-validated optimum and the validated sample is retained for future updates.
For the CFD case, $\eta_{\mathrm{phys}}$ is denoted by $\eta_{\mathrm{CFD}}$ and is set to $0.025$.

When $e_{\mathrm{phys}}>\eta_{\mathrm{phys}}$ or $\Delta J_{\mathrm{val}}\leq 0$, uncertainty-gated online refinement is activated.
A local enrichment region $\Omega_{\mathrm{enr}}(\bm{\theta}_{\mathrm{sur}}^{*})\subseteq\Omega(\mathcal{C})$ is constructed around the current surrogate optimum by shrinking the original deformation bounds with a refinement ratio $\alpha_{\mathrm{ref}}$ and clipping the resulting bounds to the originaladmissible space.
This region is used only to generate local correction samples and does not redefine theadmissible optimization space.

Additional samples are generated as
\begin{equation}
G_{\mathrm{enr}}^{(j)}
=
\mathcal{D}
\left(
G;\mathcal{C},\bm{\theta}_{\mathrm{enr}}^{(j)}
\right),
\qquad
\bm{\theta}_{\mathrm{enr}}^{(j)}
\in
\Omega_{\mathrm{enr}}
\left(
\bm{\theta}_{\mathrm{sur}}^{*}
\right),
\qquad
j=1,\ldots,N_{\mathrm{enr}},
\end{equation}
and labeled by the physics solver.
Together with the validated current candidate, these samples are added to the current training database:
\begin{equation}
\mathcal{B}_{\mathrm{cur}}^{+}
=
\mathcal{B}_{\mathrm{cur}}
\cup
\left\{
\left(
G'
\left(
\bm{\theta}_{\mathrm{sur}}^{*}
\right),
J_{\mathrm{val}}
\right)
\right\}
\cup
\left\{
\left(
G_{\mathrm{enr}}^{(j)},
J_{\mathrm{phys}}
\left(
G_{\mathrm{enr}}^{(j)}
\right)
\right)
\right\}_{j=1}^{N_{\mathrm{enr}}}.
\end{equation}
The MoE-NO surrogate is fine-tuned on $\mathcal{B}_{\mathrm{cur}}^{+}$, and the optimizer is restarted over $\Omega(\mathcal{C})$.
This process is repeated until a candidate satisfies the acceptance criteria or the maximum number of refinement iterations is reached.

% ============================================================
% ------------------------------------------------------------
\section{Numerical setup}
\label{sec:setup_details}

This section describes the numerical setup used for dataset construction, CFD labeling, data partitioning, and surrogate-model training.

\subsection{Dataset construction}

The ID aerodynamic database is built from three vehicle families, namely the multi-purpose vehicle (MPV), the sport utility vehicle (SUV), and the Sedan:
\begin{equation}
    \mathcal{V}
    =
    \{
    \mathrm{MPV},
    \mathrm{SUV},
    \mathrm{Sedan}
    \}.
\end{equation}
For each family, a production-intent baseline geometry is represented by its surface nodes, edges, and faces, as defined in Eq.~\eqref{eq:geo}.
The baseline geometries are generated using Hunyuan 3D~\cite{lai2025hunyuan3d}.
Local design variants are generated by sampling admissible deformation parameters in the knowledge-constrained design space $\Omega(\mathcal{C})$ and applying the DFFD deformation operator $\mathcal{D}$.

Each vehicle family contains 150 CFD-labeled geometries.
The historical simulation database contains 450 geometry--drag pairs:
\begin{equation}
    \mathcal{B}_{\mathrm{hist}}
    =
    \mathcal{B}_{\mathrm{MPV}}
    \cup
    \mathcal{B}_{\mathrm{SUV}}
    \cup
    \mathcal{B}_{\mathrm{Sedan}},
\end{equation}
where each sample is a geometry--drag pair $\left(G^{(i)}, J_{\mathrm{phys}}(G^{(i)})\right)$ as defined in Section~\ref{subsec:general_problem_formulation}, tagged with a vehicle-family label $z^{(i)}\in\mathcal{V}$ used for the per-family dataset splits.

An additional dataset, denoted by $\mathrm{Sedan}_{\mathrm{test}}$, is generated from another sedan-type baseline geometry that is not included in the historical simulation database. 
This geometry is first used only for distribution-level assessment. 
After it is detected as OOD, 150 local CFD-labeled variants are constructed around it using the same constrained-deformation and CFD-labeling procedure.
These 150 OOD samples are split into 120 adaptation samples, 15 validation samples, and 15 held-out test samples.
The initial MoE-NO surrogate is trained only on the MPV, SUV, and Sedan families, without using any $\mathrm{Sedan}_{\mathrm{test}}$ samples.
The $\mathrm{Sedan}_{\mathrm{test}}$ adaptation samples are introduced only after the distribution-level gate is triggered.

\begin{figure}
\centering
\includegraphics[width=0.95\textwidth]{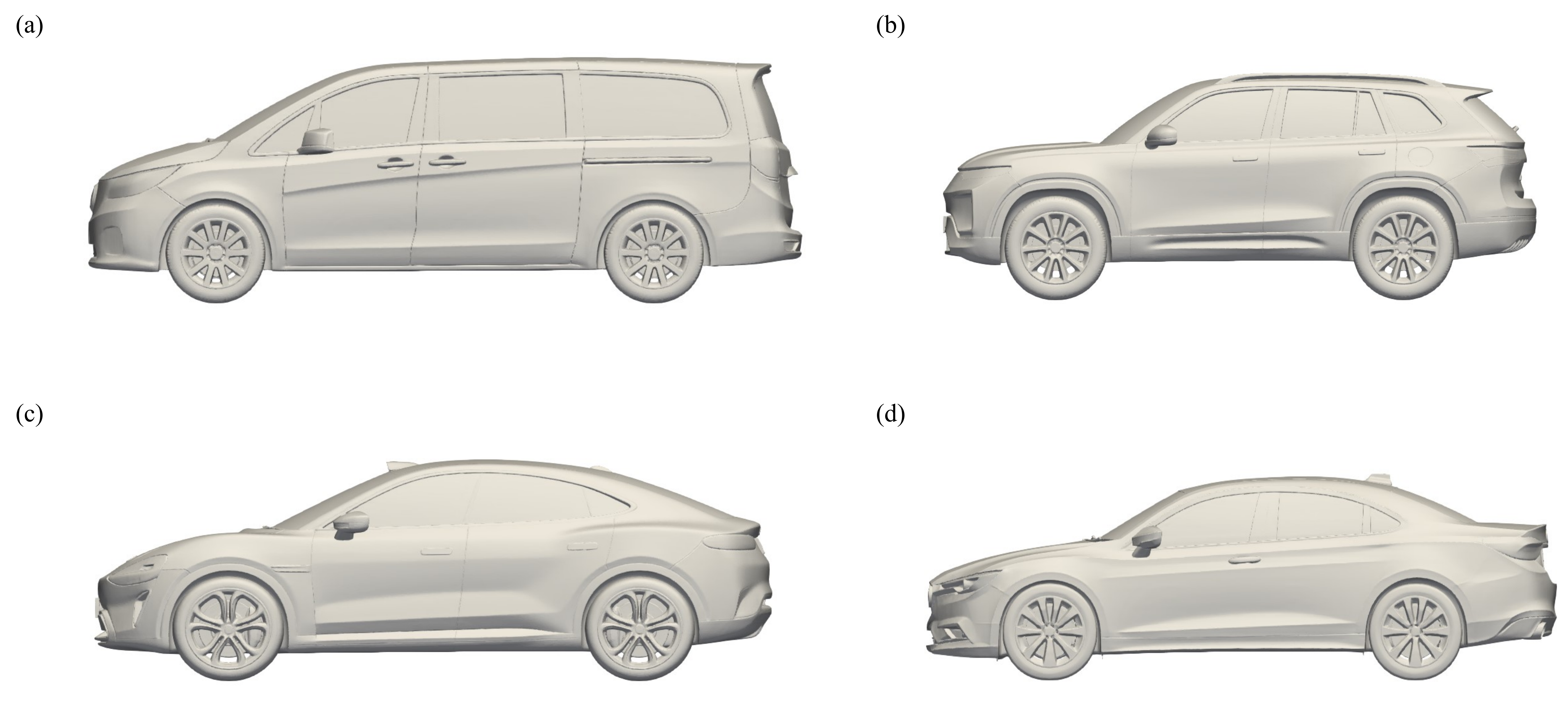}
\caption{Baseline vehicle geometries used in this study: (a) MPV, (b) SUV, (c) Sedan, and (d) the held-out $\mathrm{Sedan}_{\mathrm{test}}$ geometry. The first three vehicle families constitute the in-distribution aerodynamic database, whereas $\mathrm{Sedan}_{\mathrm{test}}$ serves as a held-out OOD test geometry.}
\label{fig:baseline_vehicle}
\end{figure}

Figure~\ref{fig:baseline_vehicle} compares the MPV, SUV, Sedan, and $\mathrm{Sedan}_{\mathrm{test}}$ baseline geometries. 
The $\mathrm{Sedan}_{\mathrm{test}}$ geometry differs from the Sedan baseline in both global proportions and local geometric details, and is therefore used as an OOD test case.

\subsection{CFD labeling}
\label{subsubsec:cfd_simulation_setup}

All baseline, sampled, locally corrected, and optimized geometries are labeled using the same high-fidelity CFD configuration.
The optimization objective is the drag coefficient $C_d$.
For a geometry $G$, the CFD label is written as
\begin{equation}
    C_d
    =
    J_{\mathrm{phys}}\left(G\right).
\end{equation}
The drag coefficient is obtained by integrating the aerodynamic drag force over the vehicle surface and normalizing it by the dynamic pressure and reference frontal area:
\begin{equation}
    C_d
    =
    \frac{
    F_D
    }{
    \frac{1}{2}\rho U_{\infty}^{2} A_{\mathrm{ref}}
    },
\end{equation}
where $F_D$ is the drag force, $\rho$ is the air density, $U_{\infty}$ is the freestream velocity, and $A_{\mathrm{ref}}$ is the reference frontal area.
In all simulations, $U_{\infty}=33.33~\mathrm{m/s}$.
The air density and dynamic viscosity are set to $\rho=1.18415\ \mathrm{kg\cdot m^{-3}}$ and $\mu=1.7893 \times 10^{-5}\ \mathrm{Pa\cdot s}$, respectively.
The reference frontal area is computed from the projected frontal area of each baseline vehicle.
The computational domain extends $75\ \mathrm{m}$, $50\ \mathrm{m}$, and 
$30\ \mathrm{m}$ in the streamwise, lateral, and vertical directions, respectively.
The inlet boundary is prescribed by the freestream velocity, the outlet uses a pressure boundary condition, and the ground and vehicle surfaces are treated as no-slip walls.

All CFD labels are computed with \textit{TF-Lattice}, a Lattice Boltzmann Method (LBM) solver \cite{hou1996lattice,liu2024influence}.
Three nested meshes are used for the grid-convergence study.
The coarse, medium, and fine meshes have minimum grid sizes of 0.0070, 0.0035, and 0.0020~m, respectively, and contain 10.22, 52.41, and 171.41 million grid points.
The grid-convergence study is conducted on a representative sedan geometry.
As shown in Table~\ref{tab:grid_independence}, the medium mesh differs from the fine mesh by only 0.29\% in $C_d$.
Considering both computational cost and numerical accuracy, the medium-resolution mesh is used for dataset construction.
The dataset has been made publicly available at \cite{fan2026github_dataset}.

\begin{table}
\centering
\caption{Grid-convergence study for a representative sedan geometry.}
\label{tab:grid_independence}
\begin{tabular}{ccccc}
\toprule
Mesh level & Minimum grid size (m) & Total grid points (million)& $C_d$ & Difference from fine-level mesh \\
\midrule
Coarse & 0.0070 & 10.22 & 0.2728 & 2.01\% \\
Medium & 0.0035 & 52.41 &  0.2776 & 0.29\% \\
Fine & 0.0020 & 171.41 & 0.2784 & -- \\
\bottomrule
\end{tabular}
\end{table}

\subsection{Dataset splits and training}

For each ID vehicle family, the CFD-labeled samples are partitioned into training, validation, and testing subsets with a ratio of $8:1:1$:
\begin{equation}
    \mathcal{B}_{z}
    =
    \mathcal{B}_{z}^{\mathrm{train}}
    \cup
    \mathcal{B}_{z}^{\mathrm{val}}
    \cup
    \mathcal{B}_{z}^{\mathrm{test}},
    \qquad
    z \in \mathcal{V} .
\end{equation}
The $\mathrm{Sedan}_{\mathrm{test}}$ samples are excluded from the initial ID training, validation, and testing splits.
They are used only after the distribution-level distribution-level gate is triggered.
At that stage, the 150 locally generated CFD-labeled $\mathrm{Sedan}_{\mathrm{test}}$ variants are split into 120 adaptation samples, 15 validation samples, and 15 held-out test samples.
The 120 adaptation samples are used only for OOD fine-tuning, whereas the held-out test samples are used only for final OOD evaluation.
The resulting dataset composition is summarized in Table~\ref{tab:vehicle_dataset}.

\begin{table}
\centering
\caption{Dataset composition and train/validation/test partitioning used in the numerical experiments. MPV, SUV, and Sedan form the ID aerodynamic database. $\mathrm{Sedan}_{\mathrm{test}}$ is the held-out OOD family; its samples are split into a distribution-level adaptation subset (listed under Train/adapt), a validation subset, and an OOD test subset.}
\label{tab:vehicle_dataset}
\begin{tabular}{llccc}
\toprule
Dataset & Role & Train/adapt & Validation & Test \\
\midrule
MPV   & ID & 120 & 15 & 15 \\
SUV   & ID & 120 & 15 & 15 \\
Sedan & ID & 120 & 15 & 15 \\
\midrule
Total (ID) & -- & 360 & 45 & 45 \\
\midrule
$\mathrm{Sedan}_{\mathrm{test}}$ & OOD & 120 & 15 & 15 \\
\bottomrule
\end{tabular}
\end{table}

Three surrogate configurations are trained to compare different model architectures under the same database setting.
The Transolver~\cite{wu2024transolver}, DragSolver~\cite{liu2025dragsolver}, and proposed encoder-based MoE-NO surrogate are all trained on the complete ID training set composed of MPV, SUV, and Sedan samples.
For each configuration, validation uses the ID validation subsets from all three ID vehicle families,
and the held-out $\mathrm{Sedan}_{\mathrm{test}}$ test subset is used only for final OOD generalization tests after distribution-level adaptation.
The three trained models are summarized in Table~\ref{tab:model}.

\begin{table} 
\centering 
\caption{Surrogate-model configurations used in the numerical experiments. All models are trained using the ID training subsets from MPV, SUV, and Sedan samples.} 
\label{tab:model} 
\begin{tabular}{lccc} 
\toprule 
Model name & $\mathcal{B}_{\mathrm{MPV}}$ & $\mathcal{B}_{\mathrm{SUV}}$ & $\mathcal{B}_{\mathrm{Sedan}}$ \\ 
\midrule 
Transolver & $\checkmark$ & $\checkmark$ & $\checkmark$ \\ 
DragSolver & $\checkmark$ & $\checkmark$ & $\checkmark$ \\ 
MoE-NO & $\checkmark$ & $\checkmark$ & $\checkmark$ \\ 
\bottomrule 
\end{tabular} 
\end{table}

All surrogate models are trained with the same supervised regression protocol to ensure
a fair comparison across architectures. The training
objective is the mean-squared error (MSE) between the predicted drag coefficient and the
CFD-labeled value.
Model parameters are optimized using AdamW with an initial learning
rate of $1\times10^{-3}$, a weight decay of $1\times10^{-4}$, and a batch size of $8$.
The learning rate is updated using a cosine-annealing-with-warmup
(\textit{CosineAnnealingWarmupRestarts}, warmup ratio $0.1$, $\eta_{\min}=1\times10^{-8}$)
scheduler, and each model is trained for at most $200$ epochs.
The checkpoint with the lowest validation MSE is retained for final evaluation.
No test samples, including the held-out $\mathrm{Sedan}_{\mathrm{test}}$ test subset, are used for early stopping, model selection, or hyperparameter tuning.

\section{Results}
\label{sec:results}

This section evaluates the proposed shape optimization framework for vehicle aerodynamic design.
The results are organized into design-space construction, surrogate accuracy, ID/OOD detection, ID optimization, and OOD adaptation.
First, we examine whether the knowledge-constrained construction procedure produces physically meaningful editable regions and admissible DFFD deformation spaces.
The Sedan case is used to show the detailed mapping from aerodynamic knowledge to geometric control variables, whereas the MPV and SUV cases are used to assess robustness across different body styles.
Second, we evaluate the accuracy of the MoE-NO surrogate on the in-distribution aerodynamic database and compare it with baseline neural-operator models.
After that, the ID/OOD detection results are reported.
Finally, we present CFD-validated optimization results for ID and OOD cases and examine the effect of the uncertainty-gated online refinement mechanism.

\subsection{Knowledge-constrained design-space construction}
\label{subsec:knowledge_constrained_design_space}

This subsection reports the constructed editable regions and the corresponding three-dimensional control boxes.
The purpose is to verify that the knowledge-derived regions can be mapped to localized geometric variables for the subsequent DFFD-based optimization.

\begin{figure}
    \centering
    \includegraphics[width=0.95\textwidth]{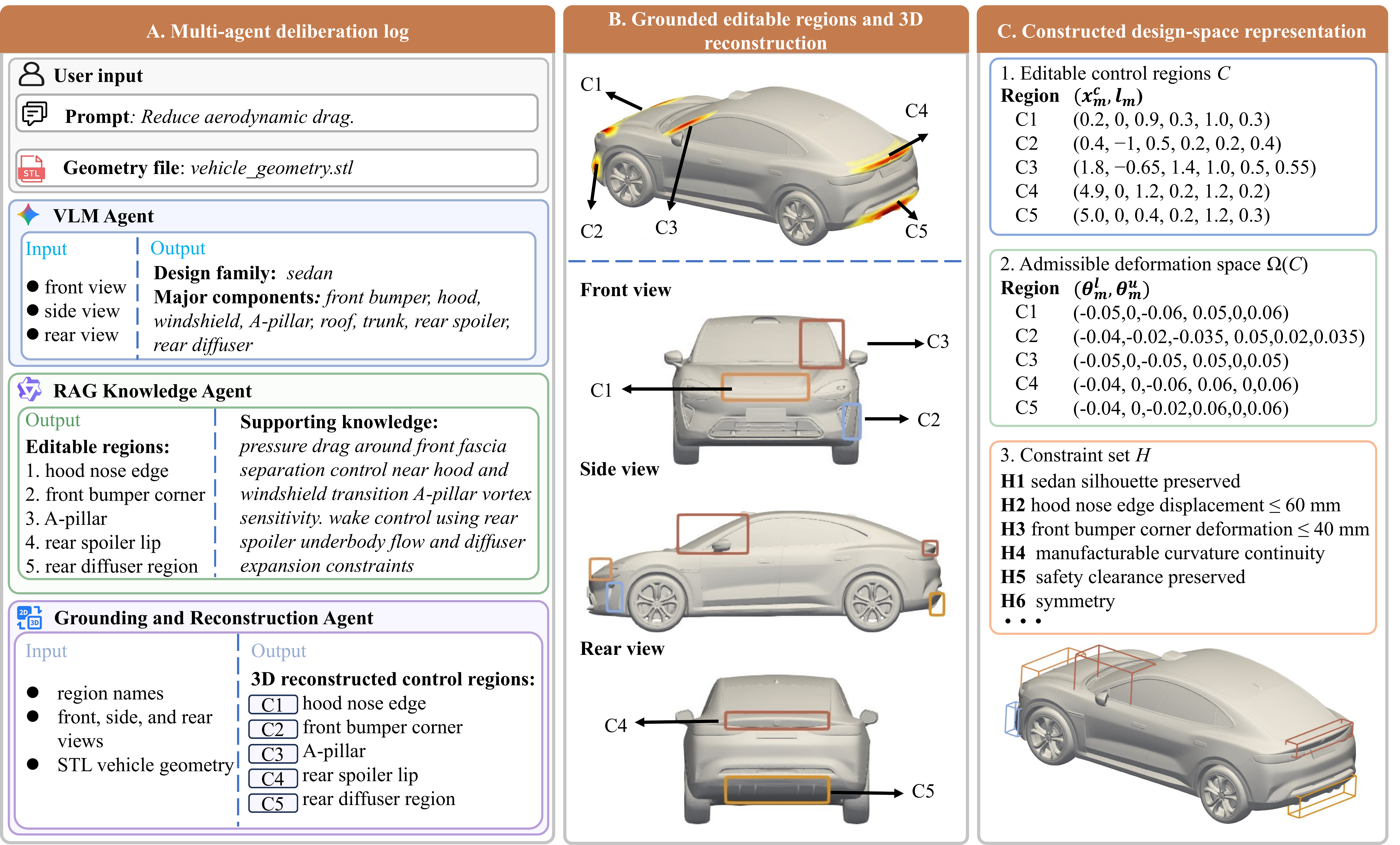}
    \caption{Knowledge-grounded construction of editable regions and constrained design-space representation. 
    (A) Multi-agent deliberation log for identifying drag-sensitive editable regions from user input, vehicle geometry, visual observations, and retrieved aerodynamic knowledge. 
    (B) Grounding of the editable regions on multi-view renderings and reconstruction of the corresponding three-dimensional control boxes. 
    (C) Constructed design-space representation, including editable control boxes $\mathcal{C}$, the admissible deformation space $\Omega(\mathcal{C})$, and preservation constraints $\mathcal{H}$.}
    \label{fig:knowledge_flow}
\end{figure}

Figure~\ref{fig:knowledge_flow} presents the intermediate outputs of the proposed pipeline.
The retrieved knowledge provides vehicle-aerodynamics rules and design-preservation constraints, the LLM converts these rules into an explicit list of editable components and admissible deformation ranges, and the VLM grounds the components on the rendered vehicle views.
The grounded image-space regions are then fused with the calibrated multi-view geometry to obtain the corresponding three-dimensional editable control boxes.
The final result is a structured design-space representation consisting of the editable control boxes $\mathcal{C}$, the admissible deformation space $\Omega(\mathcal{C})$, and the preservation constraint set $\mathcal{H}$.

In the representative Sedan case shown in Fig.~\ref{fig:knowledge_flow}, the identified editable regions include the hood nose edge, front bumper corner, A-pillar, rear spoiler lip, and rear diffuser region.
These regions are aligned with the retrieved aerodynamic knowledge: the front bumper and hood nose are related to front-fascia pressure drag and flow attachment, the A-pillar affects the windshield-side-flow transition, and the rear spoiler and diffuser regions are associated with wake control and underbody-flow expansion.
The reconstructed control boxes provide localized geometric regions for downstream deformation, while the corresponding deformation bounds and constraints restrict the optimizer to design-preserving shape changes.

The same pipeline is applied without manual case-specific modification to MPV, SUV, and Sedan geometries, as well as to the held-out $\mathrm{Sedan}_{\mathrm{test}}$ geometry used in the later out-of-distribution experiments.
The representative three-dimensional control boxes obtained for the three vehicle families are reported in Table~\ref{tab:agent_control_boxes}.

\begin{table}[htbp]
\centering
\caption{Representative three-dimensional control boxes obtained from the knowledge-constrained construction procedure. MPV, SUV, and Sedan are the in-distribution families, whereas $\mathrm{Sedan}_{\mathrm{test}}$ is the held-out OOD family used in the later out-of-distribution experiments.}
\label{tab:agent_control_boxes}
\small
\renewcommand{\arraystretch}{1.15}
\begin{tabular}{llcc}
\toprule
Vehicle type & Region & $\bm{x}_m^c$ (m) & $\bm{l}_m$ (m) \\
\midrule
MPV & Front bumper corner & $(0.35,-1.0,0.55)$ & $(0.35,0.25,0.5)$ \\
    & Hood nose edge & $(0.25,0,1.0)$ & $(0.35,1.1,0.35)$ \\
    & Roof--tailgate junction & $(5.3,0,1.9)$ & $(0.2,1.3,0.2)$ \\
    & Rear spoiler lip & $(5.4,0,1.3)$ & $(0.2,1.3,0.2)$ \\
    & Rear diffuser region & $(5.5,0,0.45)$ & $(0.2,1.3,0.35)$ \\
\midrule
SUV & Front bumper corner & $(0.3,-1.1,0.8)$ & $(0.3,0.2,0.3)$ \\
    & Hood nose edge & $(0.2,0,1.2)$ & $(0.2,1.2,0.2)$ \\
    & Rear windshield--trunk & $(4.7,0,1.8)$ & $(0.3,1.2,0.2)$ \\
    & Trunk trailing edge & $(5.1,0,1.3)$ & $(0.2,1.2,0.2)$ \\
    & Diffuser/underbody & $(5.1,0,0.5)$ & $(0.2,1.2,0.2)$ \\
\midrule
Sedan & Hood nose edge & $(0.2,0,0.9)$ & $(0.3,1.0,0.3)$ \\ 
      & Front bumper corner & $(0.4,-1,0.5)$ & $(0.2,0.2,0.4)$ \\
      & A-pillar & $(1.8,-0.65,1.4)$ & $(1.0,0.5,0.55)$ \\
      & Rear spoiler lip & $(4.9,0,1.2)$ & $(0.2,1.2,0.2)$ \\
      & Rear diffuser region & $(5.0,0,0.4)$ & $(0.2,1.2,0.3)$ \\
\midrule
$\mathrm{Sedan}_{\mathrm{test}}$
      & Hood nose edge & $(0.15,0,0.8)$ & $(0.2,0.6,0.2)$ \\ 
      & Front bumper corner & $(0.4,-0.9,0.4)$ & $(0.3,0.2,0.3)$ \\
      & A-pillar & $(1.8,-0.7,1.2)$ & $(0.8,0.4,0.5)$ \\
      & Rear spoiler lip & $(5.0,0,1.1)$ & $(0.2,1.0,0.2)$ \\
      & Rear diffuser region & $(5.1,0,0.3)$ & $(0.2,1.0,0.2)$ \\
\bottomrule
\end{tabular}
\end{table}

\subsection{MoE-NO accuracy and uncertainty quantification}
\label{subsec:moe_surrogate_accuracy}

We compare the proposed MoE-NO, denoted as MoE-NO hereafter, with two baseline models: Transolver and DragSolver.
Following the setup given in Sec.~\ref{sec:setup_details}, the comparison is performed on the MPV--SUV--Sedan multi-distribution database, and the same train--validation--test split is used for all models.

Table~\ref{tab:surrogate_comparison} summarizes the quantitative ID regression performance on the training, validation, and test subsets.
The table reports $R^2$, mean absolute error (MAE), mean absolute percentage error (MAPE), the percentage of samples with $\mathrm{MAPE}>1\%$, and the trend accuracy $\mathrm{Acc}_{\mathrm{tre}}$.
$R^2$, MAE, and MAPE evaluate pointwise regression accuracy, while the percentage of high-MAPE samples measures the frequency of relatively large prediction errors.
The trend accuracy evaluates whether the surrogate preserves the relative performance ranking among different candidate geometries, which is important for optimization.
The pointwise error metrics are defined as:
\begin{align}
    R^2
    &=
    1 -
    \frac{
    \sum_{i=1}^{N}
    \left(y_i-\hat{y}_i\right)^2
    }{
    \sum_{i=1}^{N}
    \left(y_i-\bar{y}\right)^2
    },
    \\
    \mathrm{MAE}
    &=
    \frac{1}{N}
    \sum_{i=1}^{N}
    \left|y_i-\hat{y}_i\right|,
    \\
    \mathrm{MAPE}
    &=
    \frac{100\%}{N}
    \sum_{i=1}^{N}
    \left|
    \frac{y_i-\hat{y}_i}{y_i}
    \right|,
\end{align}
where $N$ is the number of samples, $y_i$ and $\hat{y}_i$ denote the CFD-labeled and predicted drag coefficients of the $i$-th sample, respectively, and $\bar{y}$ is the mean value of all CFD-labeled drag coefficients.
The trend accuracy is defined as:
\begin{equation}
\mathrm{Acc}_{\mathrm{tre}}
=
\frac{2}{N(N-1)}
\sum_{i=1}^{N-1}
\sum_{j=i+1}^{N}
\mathbb{I}
\left[
(\hat{y}_{j}-\hat{y}_{i})(y_{j}-y_{i})>0
\right],
\end{equation}
where $\mathbb{I}[\cdot]$ is the indicator function.
Thus, $\mathrm{Acc}_{\mathrm{tre}}$ measures the proportion of design pairs for which the surrogate model correctly predicts the variation direction of the response.

\begin{table}[htbp]
\centering
\caption{ID regression performance on the historical 450-sample database.}
\label{tab:surrogate_comparison}
\small
\renewcommand{\arraystretch}{1.15}
\setlength{\tabcolsep}{5pt}
\begin{tabular}{llccccc}
\toprule
Dataset & Model & $R^2 \uparrow$ & MAE $\downarrow$ & MAPE $\downarrow$ & Percent of $\mathrm{MAPE}>1\%$ $\downarrow$ & $\mathrm{Acc}_{\mathrm{tre}}$ $\uparrow$\\
\midrule
Train & Transolver & 0.9242 & 0.0052 & 1.58\% & 40.30\% & 0.9442 \\
      & DragSolver & 0.9697 & 0.0033 & 0.98\% & 57.20\% & 0.8941 \\
      & MoE-NO & \textbf{0.9918} & \textbf{0.0016} & \textbf{0.46\%} & \textbf{9.42\%} & \textbf{0.9769}\\
\midrule
Validation & Transolver & 0.9040 & 0.0047 & 1.43\% & 48.93\% & 0.8909 \\
           & DragSolver & 0.9448 & 0.0038 & 1.15\% & 48.92\% & 0.8303 \\
           & MoE-NO & \textbf{0.9510} & \textbf{0.0036} & \textbf{1.08\%} & \textbf{43.96\%} & \textbf{0.8990} \\
\midrule
Test & Transolver & 0.9007 & 0.0057 & 1.58\% & 60.11\% & 0.9051 \\
     & DragSolver & 0.9381 & 0.0049 & 1.52\% & 60.05\% & 0.8495 \\
     & MoE-NO & \textbf{0.9503} & \textbf{0.0038} & \textbf{1.16\%} & \textbf{33.31\%} & \textbf{0.9434}\\
\bottomrule
\end{tabular}
\end{table}

The results show that MoE-NO achieves the best performance across all three subsets and all reported metrics.
On the training subset, MoE-NO reaches $R^2=0.9918$, MAE $=0.0016$, MAPE $=0.46\%$, and $\mathrm{Acc}_{\mathrm{tre}}=0.9769$, outperforming both Transolver and DragSolver.
On the validation subset, MoE-NO also gives the highest $R^2$ of $0.9510$, the lowest MAE of $0.0036$, the lowest MAPE of $1.08\%$, and the highest trend accuracy of $0.8990$.
On the test subset, MoE-NO maintains the best generalization performance, with $R^2=0.9503$, MAE $=0.0038$, MAPE $=1.16\%$, and $\mathrm{Acc}_{\mathrm{tre}}=0.9434$.
The percentage of test cases with $\mathrm{MAPE}>1\%$ is reduced to $33.31\%$ for MoE-NO, compared with $60.11\%$ for Transolver and $60.05\%$ for DragSolver.

\begin{figure}
    \centering
    \includegraphics[width=0.99\textwidth]{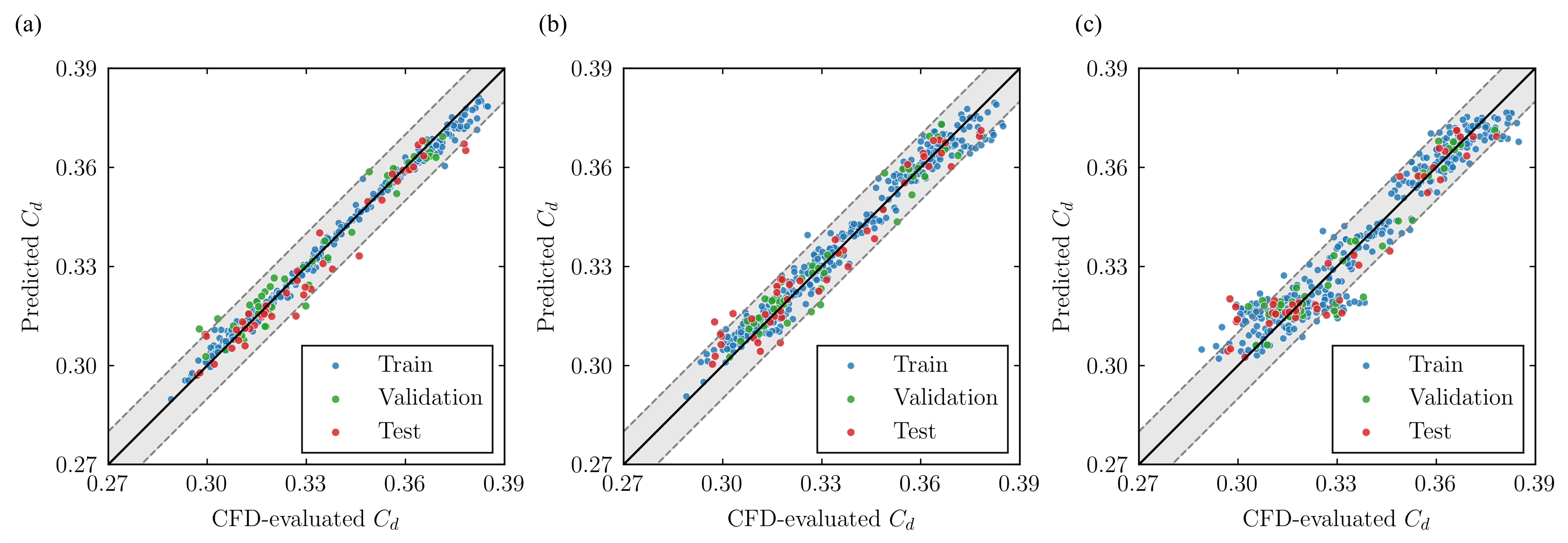}
    \caption{Prediction comparison of different surrogate models. (a) MoE-NO. (b) DragSolver. (c) Transolver.}
    \label{fig:moe_results}
\end{figure}

Figure~\ref{fig:moe_results} further visualizes the test-set prediction accuracy by comparing the predicted $C_d$ with the CFD-labeled values.
Most MoE-NO predictions are located inside or close to the $\pm1\%$ error band, whereas Transolver and DragSolver show larger deviations in several local regions.

\begin{figure}
    \centering
    \includegraphics[width=0.6\textwidth]{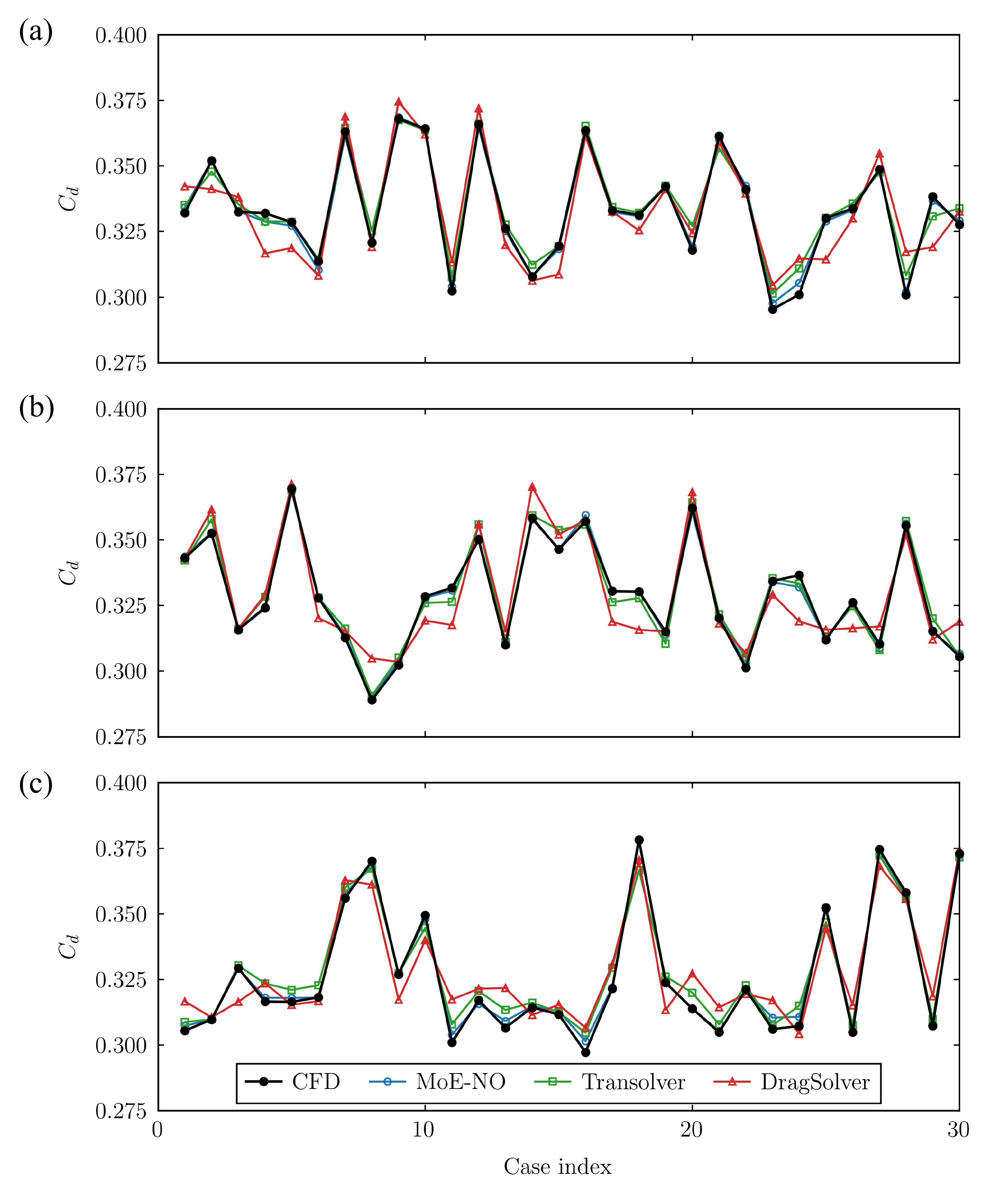}
    \caption{Comparison of the $C_d$ variation trend predicted by MoE-NO, Transolver, and DragSolver on 30 randomly selected test cases from (a) MPV (b) SUV (c) Sedan. The CFD-labeled results are used as the reference.}
    \label{fig:acc_trend}
\end{figure}

After the quantitative comparison, Fig.~\ref{fig:acc_trend} qualitatively examines whether the models can follow the variation trend of $C_d$ across 30 randomly selected test cases from MPV, SUV, and Sedan.
MoE-NO follows the local increase and decrease of the CFD-labeled drag coefficient more closely than the two baseline models.
This behavior is consistent with the higher test-set trend accuracy reported in Table~\ref{tab:surrogate_comparison} and indicates that MoE-NO is more reliable for ranking candidate geometries during optimization.

The encoder-based MoE-NO also provides the latent features used for reliability estimation.
We use the relative prediction error to examine whether the percentile uncertainty score is correlated with surrogate error:
\begin{equation}
\delta
=
\left|
\frac{
C_d^{\mathrm{CFD}} - C_d^{\mathrm{NO}}
}{
C_d^{\mathrm{CFD}}
}
\right| \times 100\%,
\end{equation}
where $C_d^{\mathrm{CFD}}$ denotes the CFD-labeled drag coefficient, i.e., the test-set instantiation of the physics objective $J_{\mathrm{phys}}$ defined in Section~\ref{subsec:general_problem_formulation}, and $C_d^{\mathrm{NO}}=\hat{J}_{\phi}$ denotes the drag coefficient predicted by the neural operator.
Since the percentile uncertainty $\sigma$ defined in Eq.~\eqref{eq:mahalanobis_percentile} already lies in $[0,1]$, it is used directly for visualization.

\begin{figure}
    \centering
    \includegraphics[width=0.45\textwidth]{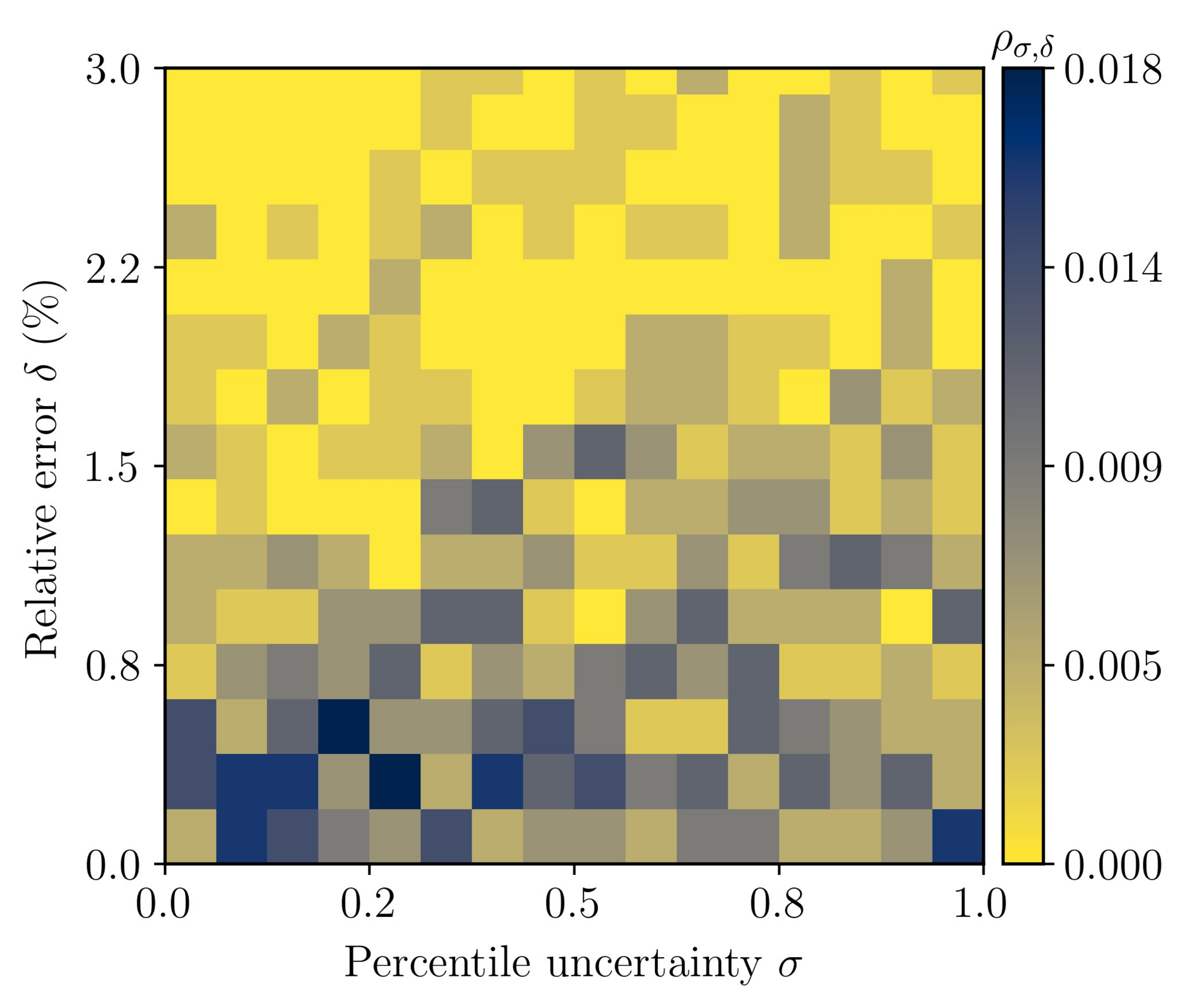}
    \caption{Joint distribution between percentile uncertainty $\sigma$ and relative error $\delta$ on the test dataset. The color denotes the estimated probability density $\rho_{\sigma,\delta}$ of samples in each bin.}
    \label{fig:uncertainty}
\end{figure}

Fig.~\ref{fig:uncertainty} shows the joint probability distribution density $\rho_{\sigma,\delta}$ between the percentile uncertainty $\sigma$ and the relative error $\delta$ on all test samples.
The horizontal axis is the percentile uncertainty $\sigma$, the vertical axis is the relative error $\delta$, and the color denotes the estimated probability density $\rho_{\sigma,\delta}$ of samples in each bin.
The probability mass in the upper-left region, where the prediction error is large but the estimated uncertainty is low, is very small.
This indicates that the uncertainty indicator has a low probability of missing high-error cases, which is the most critical failure mode for an uncertainty-gated optimization loop.
Although some low-error samples are assigned relatively large uncertainty values, this behavior corresponds to conservative verification rather than unsafe acceptance.
From the perspective of surrogate-assisted optimization, such conservative uncertainty reporting is preferable because it may trigger additional solver checks for reliable candidates, but it reduces the risk of accepting an inaccurate surrogate optimum without physical validation.

\subsection{OOD detection}
\label{subsec:id_ood_detection}

\begin{figure}
    \centering
    \includegraphics[width=0.5\textwidth]{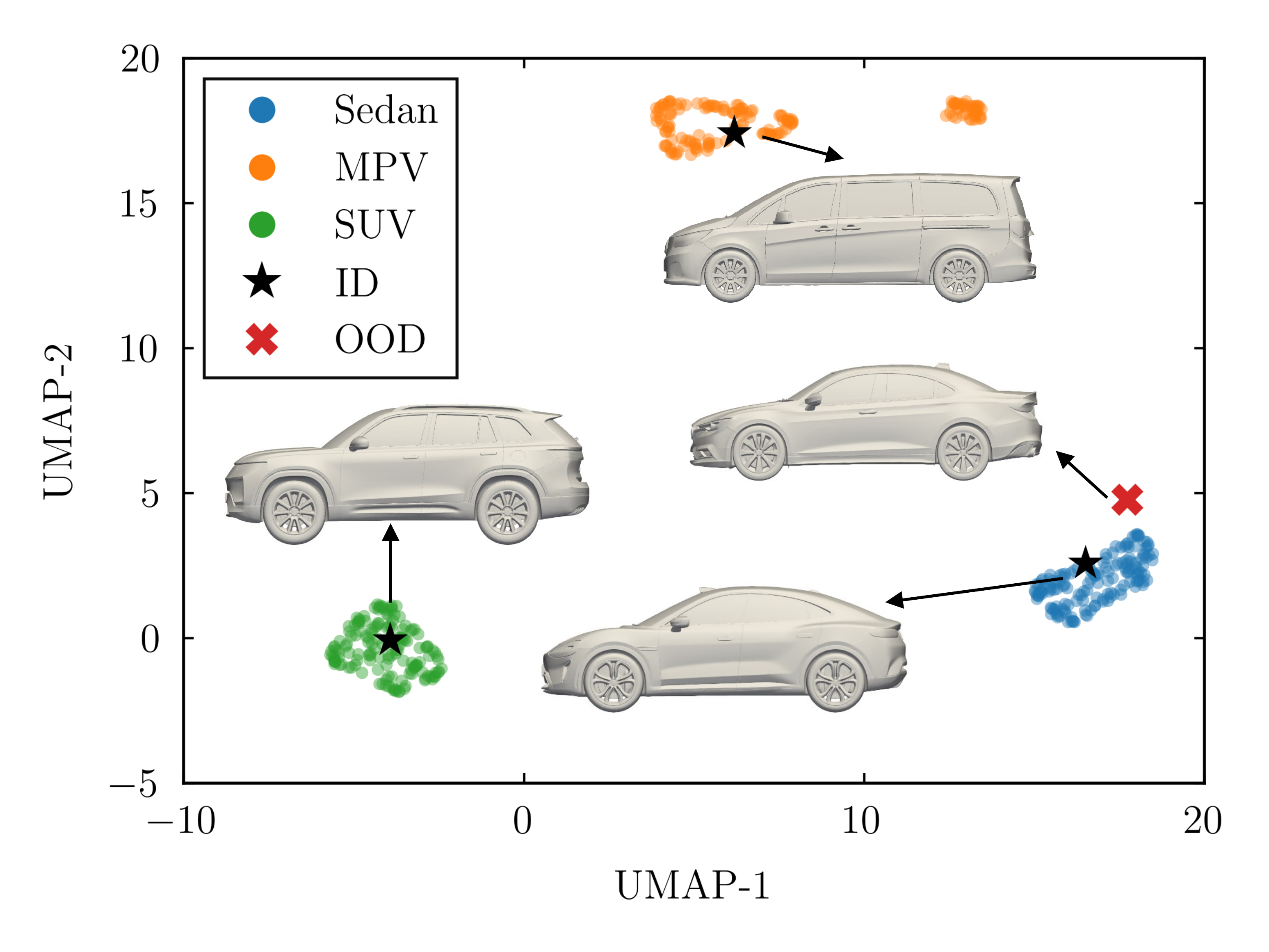}
    \caption{UMAP visualization of the encoder embeddings. The ID training samples form three clusters corresponding to MPV, SUV, and Sedan geometries, whereas the $\mathrm{Sedan}_{\mathrm{test}}$ samples are located outside the training clusters.}
    \label{fig:case_retrieval_results}
\end{figure}

After the MoE-NO surrogate is trained, its encoder provides the latent representation used for database-coverage assessment.
Figure~\ref{fig:case_retrieval_results} visualizes the training samples and query samples by projecting the learned embeddings onto a two-dimensional plane using Uniform Manifold Approximation and Projection (UMAP)~\cite{mcinnes2018umap}.
The ID training samples form three separated clusters corresponding to the MPV, SUV, and Sedan families.
The $\mathrm{Sedan}_{\mathrm{test}}$ samples are located outside these clusters, indicating that this vehicle family is not covered by the original aerodynamic database.

\begin{table}[htbp]
\centering
\caption{Representative encoder-space metrics for an ID Sedan query and an OOD $\mathrm{Sedan}_{\mathrm{test}}$ query.}
\label{tab:retrieval_results}
\begin{tabular}{lccccccc}
\toprule
Query & $d_{\mathrm{Euc}}$ & $d_{\mathrm{KNN}}$ & $s_{\mathrm{cos}}$ & $I_{\mathrm{Euc}}$ & $I_{\mathrm{KNN}}$ & $I_{\mathrm{cos}}$ & Decision \\
\midrule
Sedan & 35.107 & 36.799 & 0.987332 & Pass & Pass & Pass & ID \\
$\mathrm{Sedan}_{\mathrm{test}}$ & 49.163 & 50.591 & 0.974234 & Fail & Fail & Fail & OOD \\
\bottomrule
\end{tabular}
\end{table}

Table~\ref{tab:retrieval_results} reports representative quantitative metrics.
The ID Sedan query remains close to the training database in both nearest-neighbor and local-neighborhood distances and also has a high cosine similarity.
In contrast, the $\mathrm{Sedan}_{\mathrm{test}}$ query has larger encoder-space distances and it fails all three criteria.
Therefore, $\mathrm{Sedan}_{\mathrm{test}}$ is treated as an OOD vehicle family in the following optimization experiments.

\subsection{Uncertainty-gated online refinement}
\label{subsec:cfd_validated_optimization}

This subsection evaluates the online optimization process for the MPV, SUV, and Sedan baseline geometries.
For each case, the optimizer searches within the knowledge-constrained DFFD design space defined by the control boxes in Table~\ref{tab:agent_control_boxes}.
The purpose is to examine whether the surrogate-guided optimizer can identify low-drag designs and whether the uncertainty estimate can indicate when additional CFD validation and online refinement are required.

\begin{figure}
    \centering
    \includegraphics[width=0.99\textwidth]{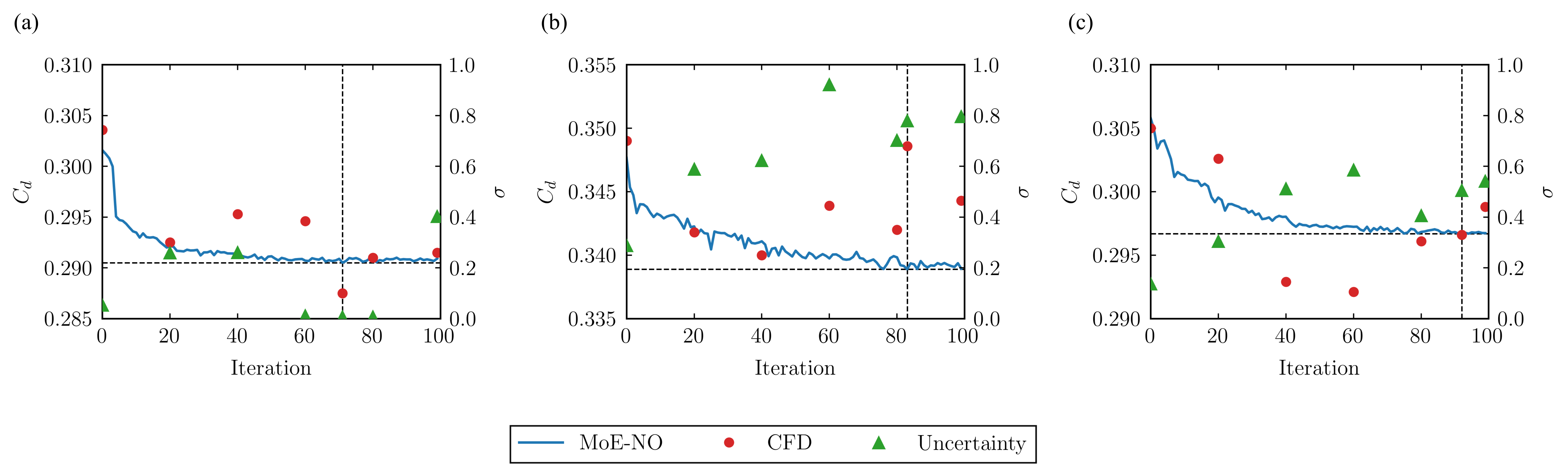}
    \caption{Optimization traces without online refinement for (a) MPV (b) SUV (c) Sedan.}
    \label{fig:optimization_trace_without_refinement}
\end{figure}

Figure~\ref{fig:optimization_trace_without_refinement} shows the optimization traces obtained before online refinement.
The blue curve denotes the surrogate-predicted $C_d$ over the optimization iterations, the red markers denote diagnostic CFD-validated checkpoints, and the green markers denote the Mahalanobis-percentile uncertainty $\sigma$.
For diagnostic analysis, CFD evaluations are additionally performed every 20 generations and at the surrogate-predicted optimum.
These checkpoints are not used as optimization-decision gates; online refinement is still governed by $\eta_{\sigma}$ and $\eta_{\mathrm{CFD}}$.
The horizontal dashed line marks the CFD-validated $C_d$ of the selected candidate, and the vertical dashed line marks the corresponding iteration.
In all three cases, the surrogate-predicted drag coefficient decreases as the search proceeds, indicating that the optimizer can locate promising low-drag candidates within the constrained design space.
However, because the search is driven by an approximate surrogate, the final design cannot be accepted solely according to the predicted $C_d$.

The uncertainty-gated decision rule is therefore used to determine whether the surrogate optimum should be accepted directly or sent to the high-fidelity CFD solver for verification.
If $\sigma \leq \eta_{\sigma}$, the candidate is considered to lie in a sufficiently supported region of the reference latent distribution and no online refinement is activated.
For reporting the final performance, a CFD evaluation is still conducted for the accepted design.
If $\sigma>\eta_{\sigma}$, the candidate is first verified by CFD, and online refinement is activated only when the verified surrogate--CFD discrepancy exceeds $\eta_{\mathrm{CFD}}$ or when CFD does not confirm the surrogate-predicted drag reduction.

The checkpoint CFD results in Fig.~\ref{fig:optimization_trace_without_refinement} show how the uncertainty signal is used to decide whether the surrogate optimum requires CFD verification and refinement. 
For MPV and Sedan, the selected optima remain in relatively low-uncertainty regions, and the CFD-validated results are consistent with the surrogate-predicted improvements.
These two cases are therefore accepted without online refinement.
For SUV, the surrogate-predicted optimum is associated with a larger uncertainty value and a larger CFD discrepancy.
Specifically, the surrogate-only optimum predicts $C_d=0.3389$, whereas CFD validation gives $C_d=0.3486$, corresponding to a surrogate--CFD discrepancy of $2.78\%$.
This indicates that the surrogate is less reliable near the predicted SUV optimum, and the online refinement mechanism is consequently activated.

\begin{figure}
    \centering
    \includegraphics[width=0.95\textwidth]{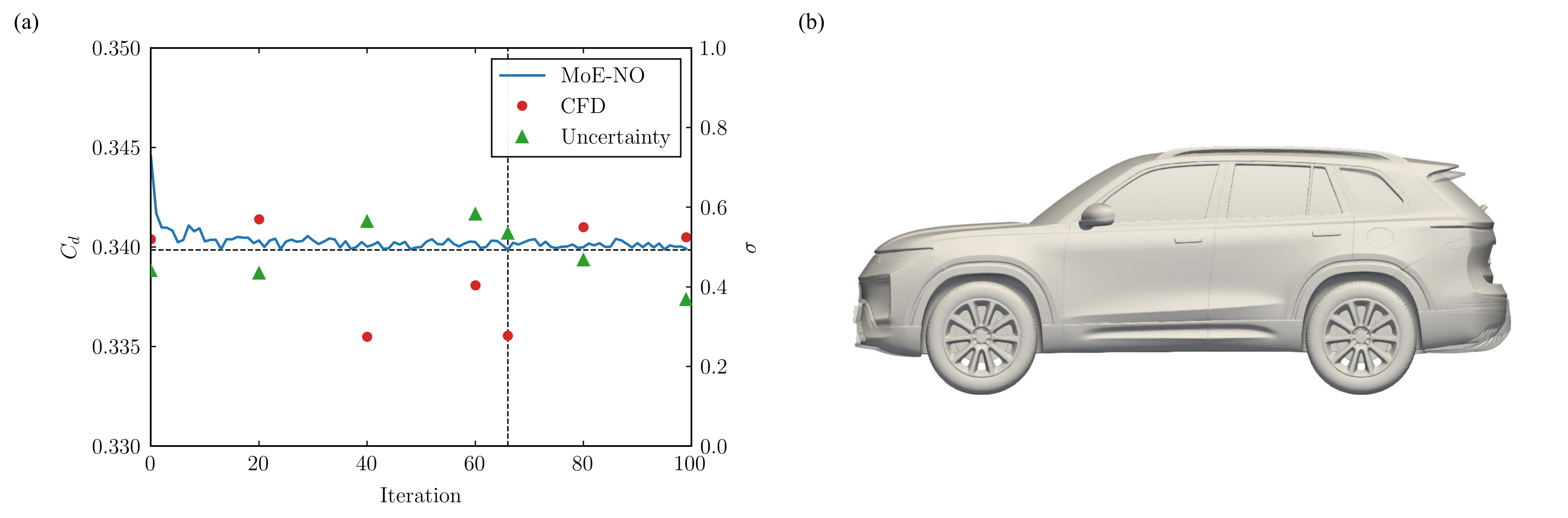}
    \caption{Online-refinement result for the SUV case. 
    (a) Optimization trace. 
    (b) Optimized geometry.}
    \label{fig:suv_online_refinement}
\end{figure}

Figure~\ref{fig:suv_online_refinement} shows the optimization result after online refinement for the SUV case.
Additional CFD-labeled samples are generated around the uncertain candidate, and the MoE-NO is fine-tuned with the augmented local dataset to correct the local response surface. The surrogate-assisted optimizer is then restarted over the original admissible deformation space $\Omega(\mathcal{C})$.
After refinement, the surrogate prediction becomes more consistent with the CFD validation.
The CFD-validated drag coefficient decreases from $0.3486$ to $0.3355$, and the surrogate--CFD discrepancy is reduced from $2.78\%$ to $1.25\%$.
This result demonstrates that uncertainty-gated online refinement improves the reliability of surrogate-guided optimization near uncertain optima.

\begin{figure}
    \centering
    \includegraphics[width=0.95\textwidth]{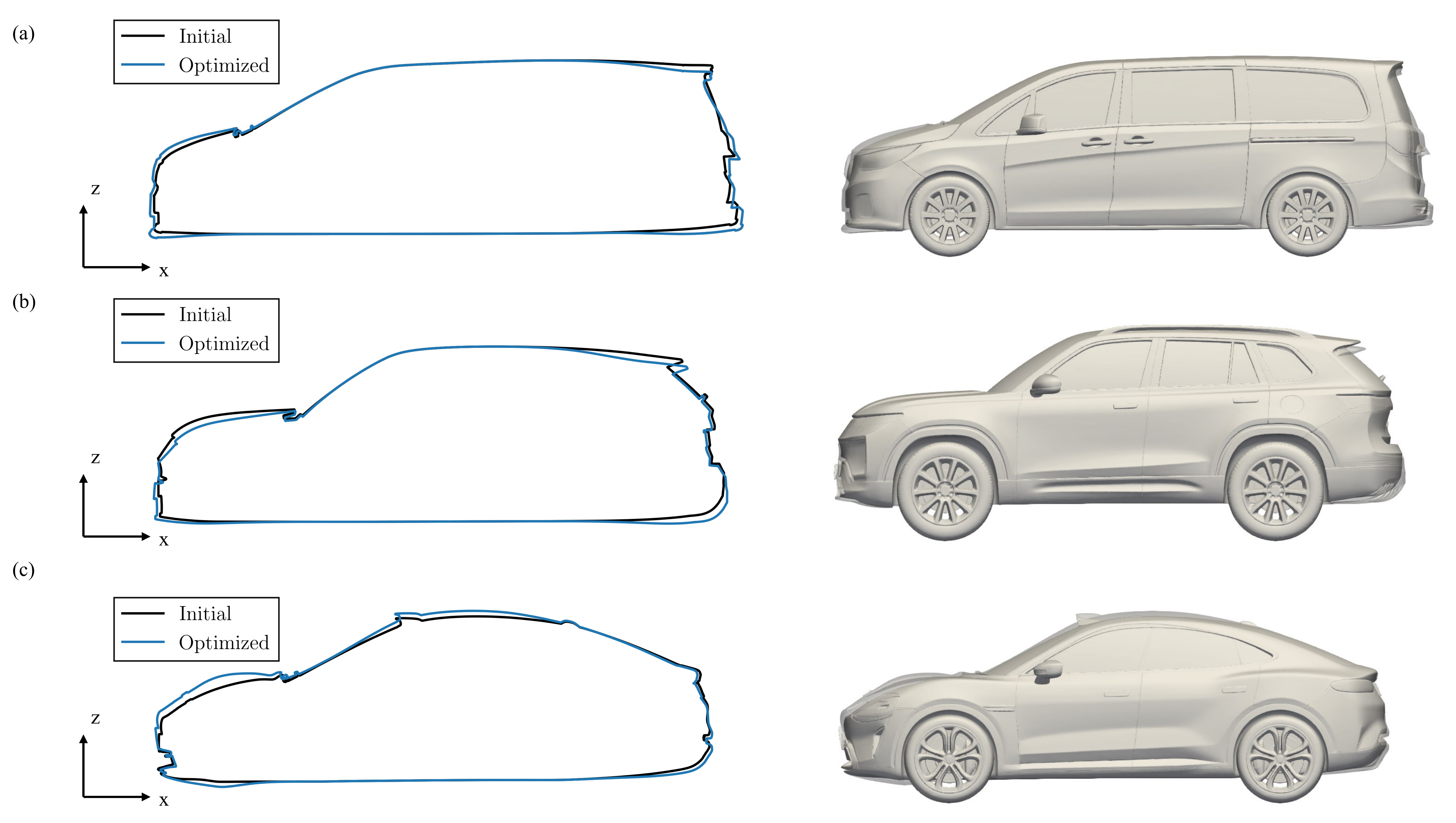}
    \caption{Comparison between the initial and optimized geometries for (a) MPV (b) SUV (c) Sedan.}
    \label{fig:optimized_geometry_comparison}
\end{figure}

The comparison between the baseline and optimized geometries is shown in Fig.~\ref{fig:optimized_geometry_comparison}.
The black contour denotes the initial geometry, and the blue contour denotes the optimized geometry.
The optimized geometries mainly modify the predefined aerodynamic-sensitive local regions while preserving the overall side-view profile and the main vehicle proportions.

\begin{table}[htbp]
\centering
\caption{CFD-validated optimization results and effect of online refinement.}
\label{tab:opt_result}
\small
\renewcommand{\arraystretch}{1.15}
\setlength{\tabcolsep}{4pt}
\begin{tabular}{ccccccc}
\toprule
Vehicle
& \shortstack{Online\\refinement}
& \shortstack{Baseline\\CFD $C_d$}
& \multicolumn{3}{c}{Selected optimum}
& \shortstack{CFD-validated\\drag reduction}\\
\cmidrule(lr){4-6}
&
&
& \shortstack{Surrogate\\predicted $C_d$}
& \shortstack{CFD\\validated $C_d$}
& \shortstack{Surrogate--CFD\\discrepancy}
& \\
\midrule
MPV
& No
& 0.3213
& 0.2905
& 0.2875
& $1.04\%$
& $10.52\%$\\
\midrule
SUV
& No
& 0.3645
& 0.3389
& 0.3486
& $2.78\%$
& $4.36\%$\\
SUV
& Yes
& 0.3645
& 0.3397
& 0.3355
& $1.25\%$
& $7.96\%$\\
\midrule
Sedan
& No
& 0.3096
& 0.2966
& 0.2967
& $0.03\%$
& $4.17\%$\\
\bottomrule
\end{tabular}
\end{table}

Table~\ref{tab:opt_result} summarizes the CFD-validated optimization and online refinement ablation results.
For each vehicle type, the table reports whether online refinement was applied, the baseline CFD drag coefficient, the surrogate-predicted optimum, the CFD-validated drag coefficient, the surrogate--CFD discrepancy, and the achieved drag reduction.
For MPV and Sedan, the surrogate-predicted optima are consistent with the CFD validation, and the final CFD results show drag reductions of $10.52\%$ and $4.17\%$, respectively.
For SUV, the surrogate-only optimization gives a lower predicted $C_d$, but the CFD validation shows a larger surrogate--CFD discrepancy of $2.78\%$.
After online refinement, the CFD-validated drag reduction increases from $4.36\%$ to $7.96\%$, and the surrogate--CFD discrepancy decreases from $2.78\%$ to $1.25\%$.
This comparison forms an ablation between the original surrogate-only optimization and the CFD-in-the-loop online refinement strategy.

\subsection{Out-of-distribution adaptation and optimization}
\label{subsec:ood_adaptation}

This subsection evaluates the distribution-level gate, which adapts the surrogate when a query geometry is detected as out-of-distribution. 
The held-out $\mathrm{Sedan}_{\mathrm{test}}$ family, classified as OOD in Section~\ref{subsec:id_ood_detection}, is used as the test case. 
Following the first physics-solver-in-the-loop mechanism, the distribution-level gate triggers the construction of $N_s=150$ local geometries around the $\mathrm{Sedan}_{\mathrm{test}}$ baseline within the expanded sampling space $\Omega_{\mathrm{sam}}^{\mathrm{ex}}(\mathcal{C})$ (a slight widening of $\Omega(\mathcal{C})$ used only for surrogate enrichment). 
All 150 geometries are labeled by CFD and then split into 120 adaptation samples, 15 validation samples, and 15 held-out test samples. 
Only the 120 adaptation samples are used to fine-tune the MoE-NO surrogate together with an additional expert $E_{N_{\mathrm{exp}}+1}$, while the held-out test samples are used only for final OOD evaluation.

Table~\ref{tab:ood_accuracy} reports the surrogate accuracy on the held-out 15-sample $\mathrm{Sedan}_{\mathrm{test}}$ test split before and after adaptation. 
Without adaptation, the surrogate trained only on the in-distribution families generalizes poorly to the OOD geometry, with $R^2=-18.42$ and $\mathrm{MAPE}=14.88\%$. 
After adding the new expert and fine-tuning on the local CFD-labeled samples, the accuracy improves to $R^2=0.9525$ and $\mathrm{MAPE}=1.84\%$, confirming that the distribution-level adaptation restores reliable prediction on the newly
encountered design family.

\begin{table}[htbp]
\centering
\caption{Out-of-distribution surrogate accuracy on the held-out 15-sample $\mathrm{Sedan}_{\mathrm{test}}$ test split,
before and after adaptation.}
\label{tab:ood_accuracy}
\small
\renewcommand{\arraystretch}{1.15}
\setlength{\tabcolsep}{5pt}
\begin{tabular}{lccccc}
\toprule
Model & $R^2 \uparrow$ & MAE $\downarrow$ & MAPE $\downarrow$ & Percent of $\mathrm{MAPE}>1\%$ $\downarrow$ & $\mathrm{Acc}_{\mathrm{tre}}$ $\uparrow$ \\
\midrule
MoE-NO (no adaptation)          & $-18.42$ & 0.0406 & 14.88\% & 100.00\% & 0.6000 \\
MoE-NO (adapted) & \textbf{0.9525} & \textbf{0.0057} & \textbf{1.84\%} & \textbf{66.67\%} & \textbf{0.9238} \\
\bottomrule
\end{tabular}
\end{table}

\begin{figure}
    \centering
    \includegraphics[width=0.95\textwidth]{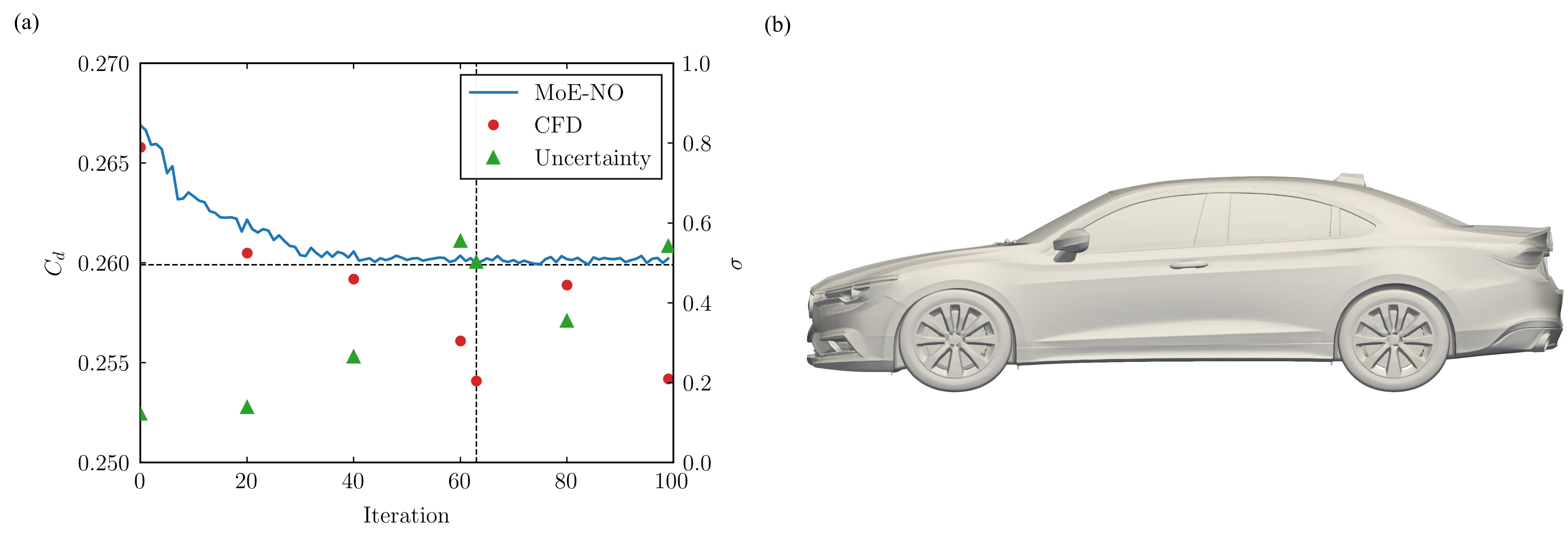}
    \caption{(a) CFD-validated optimization of the OOD $\mathrm{Sedan}_{\mathrm{test}}$ geometry obtained with the
    adapted surrogate.
    (b) Optimized $\mathrm{Sedan}_{\mathrm{test}}$ geometry.}
    \label{fig:ood_optimization}
\end{figure}

Table~\ref{tab:ood_opt} reports the CFD-validated
optimization of the $\mathrm{Sedan}_{\mathrm{test}}$ geometry obtained with the adapted surrogate. 
Starting from a baseline $C_d=0.2745$, the optimized design reaches a CFD-validated $C_d=0.2541$, a drag reduction of $7.43\%$, while the surrogate prediction at the selected optimum ($C_d=0.2599$) agrees with CFD to within a surrogate--CFD discrepancy of $2.28\%$. 
Figure~\ref{fig:ood_optimization}(a) shows the
optimization history: the surrogate-predicted $C_d$ decreases steadily over the generations and the
CFD-validated checkpoints confirm the predicted trend. The optimized geometry, shown in
Figure~\ref{fig:ood_optimization}(b), modifies the local drag-sensitive regions while preserving the
overall vehicle proportions and side-view profile. 
This demonstrates that the distribution-level gate enables reliable,
CFD-validated optimization on a design family that is not covered by the historical simulation database.

\begin{table}[htbp]
\centering
\caption{CFD-validated optimization of the OOD $\mathrm{Sedan}_{\mathrm{test}}$ case using the MoE-NO (adapted).}
\label{tab:ood_opt}
\small
\renewcommand{\arraystretch}{1.15}
\setlength{\tabcolsep}{4pt}
\begin{tabular}{ccccc}
\toprule
\shortstack{Baseline CFD $C_d$}
& \shortstack{Surrogate predicted $C_d$}
& \shortstack{Surrogate--CFD\\discrepancy}
& Prediction error
& \shortstack{CFD-validated drag reduction}\\
\midrule
0.2745 & 0.2599 & 0.2541 & $2.28\%$ & $7.43\%$\\
\bottomrule
\end{tabular}
\end{table}

\section{Conclusions}
\label{sec:conclusions}

This work developed a knowledge-constrained shape-optimization framework for high-confidence design under physics-solver evaluations.
The method combines knowledge-constrained design-space construction with a MoE-NO surrogate and uncertainty-gated solver feedback.
Engineering knowledge and user intent are converted into editable regions, deformation bounds, and design-preservation constraints, while the MoE-NO surrogate guides the search and triggers solver calls for out-of-distribution adaptation, CFD verification, or uncertainty-gated online refinement.
This makes high-fidelity solver feedback active during optimization instead of relying only on a fixed surrogate trained before the search.

Across representative automotive aerodynamic cases, including MPV, SUV, Sedan, and a held-out $\mathrm{Sedan}_{\mathrm{test}}$ geometry, the framework improves both surrogate prediction and CFD-validated optimization.
On the in-distribution MPV--SUV--Sedan database, the MoE-NO surrogate achieves higher drag-prediction accuracy and trend consistency than the Transolver and DragSolver baselines.
For the MPV and Sedan cases, the optimized geometries achieve CFD-validated drag reductions of $10.52\%$ and $4.17\%$, respectively, without requiring online refinement.
For the SUV case, uncertainty-gated refinement reduces the surrogate--CFD discrepancy from $2.78\%$ to $1.25\%$ and improves the CFD-validated drag reduction from $4.36\%$ to $7.96\%$.
For the held-out $\mathrm{Sedan}_{\mathrm{test}}$ family, distribution-level adaptation improves the OOD surrogate accuracy from $R^2=-18.42$ and $\mathrm{MAPE}=14.88\%$ to $R^2=0.9525$ and $\mathrm{MAPE}=1.84\%$, and the adapted surrogate yields a CFD-validated drag reduction of $7.43\%$.
These results indicate that database coverage and uncertainty can provide useful reliability signals for deciding when neural-operator-based optimization requires additional solver support.

Compared with fully surrogate-driven optimization, an advantage of the proposed formulation is that high-fidelity evaluations are used as selective confidence-improving corrections rather than as either a one-time data-generation step or an evaluation for every candidate design.
The distribution-level gate addresses loss of surrogate reliability caused by insufficient database coverage, while the optimization-level gate addresses extrapolation induced by the search process itself.
The knowledge-constrained construction also differs from unconstrained geometric perturbation by translating irregular engineering rules into an explicit optimization space with editable regions, admissible deformation bounds, and preservation constraints.
Together, these components provide a practical route to high-confidence design when both physical evaluation and manual design-space construction are costly.

The framework also has limitations.
It depends on the quality of the retrieved engineering knowledge, visual grounding, and reconstructed control boxes, and errors in these steps can restrict or bias the admissible deformation space.
The MoE-NO surrogate and the uncertainty gates still require a sufficiently informative historical database and additional solver-labeled samples when a new design family is poorly covered.
The present study is limited to single-objective drag reduction for vehicle aerodynamics, and the computational cost of CFD-based adaptation and online refinement remains non-negligible.
Future work will focus on multi-objective and multidisciplinary optimization, stricter manufacturability and packaging constraints, larger industrial geometry databases, and variable-fidelity solver feedback to further reduce the number of high-fidelity evaluations required for reliable shape optimization.

%\section*{Acknowledgments}

\bibliographystyle{model1-num-names}
\bibliography{refs}

\end{document}